\documentclass[10pt,twocolumn,letterpaper]{article}

\usepackage{cvpr}              %
\makeatletter\toggletrue{cvprfinal}\togglefalse{cvprrebuttal}\makeatother %

\usepackage{pifont}

\usepackage{xspace}
\newcommand{\ours}{AdProducer\xspace}

\usepackage{multirow}
\usepackage{array}
\usepackage{colortbl}
\usepackage{listings}  %
\newcommand{\oursrow}{\rowcolor{gray!12}}  %

\definecolor{cvprblue}{rgb}{0.21,0.49,0.74}
\usepackage[pagebackref,breaklinks,colorlinks,allcolors=cvprblue]{hyperref}

\def\confName{CVPR}
\def\confYear{2027}

\usepackage{fontawesome5}
\usepackage{tcolorbox}\usepackage[T1,OT1]{fontenc}  %
\tcbuselibrary{listings,breakable}  \usepackage{needspace}  %
\usepackage{textcomp}
\definecolor{base}{HTML}{B5523A}\definecolor{ours}{HTML}{3F73A8}\definecolor{prompt}{HTML}{3E7D4A}
\newcommand{\codeurl}{https://github.com/lky-ang/AdProducer}
\newcommand{\hfurl}{https://huggingface.co/AdProducer}
\newcommand{\pageurl}{https://lky-ang.github.io/AdProducer-page/}
\title{\raisebox{-0.24\height}[0.76\height][0pt]{\includegraphics[height=1.75em]{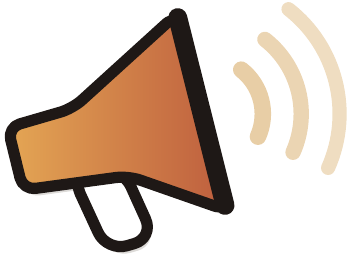}}\hspace{0.3em}\includegraphics[height=0.98em]{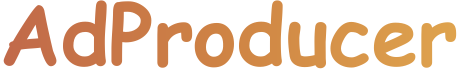}: Towards Agentic Advertising Video Generation}

\author{
Kunyang Li$^{*}$ \quad Chaoyang Wang$^{*}$ \quad Mubarak Shah \quad Yuzhang Shang\\[0.25em]
Institute of Artificial Intelligence, University of Central Florida\\[0.7em]
{\normalsize
\href{\codeurl}{\textcolor{black}{\faGithub}~Code}\hspace{2.2em}%
\href{\hfurl}{\raisebox{-0.15\height}{\includegraphics[height=1.05em]{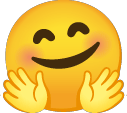}}~Models \& Benchmark}\hspace{2.2em}%
\href{\pageurl}{\textcolor{black}{\faGlobe}~Project Page}}
}

\begin{document}
\twocolumn[{%
\renewcommand\twocolumn[1][]{#1}%
\maketitle
\vspace{-0.6em}
\begin{center}
\captionsetup{type=figure}
\includegraphics[width=\textwidth]{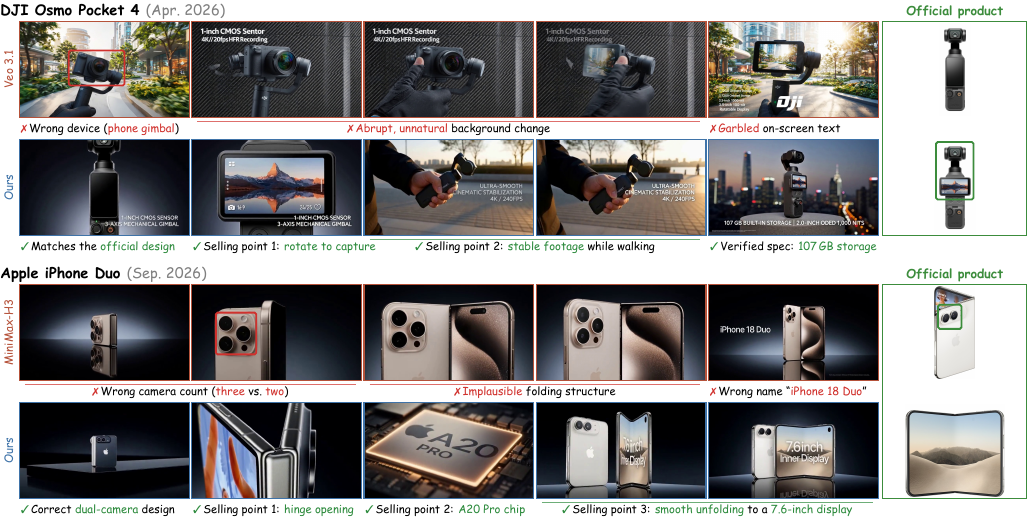}
\captionof{figure}{
\textbf{Comparison of advertising videos for products released in 2026.} Compared with direct prompting (red), AdProducer (blue) generates more faithful and compelling advertisements using product research and storyboard design with the same frozen video generator. For the DJI Osmo Pocket 4, the short brief is expanded into a shot-level storyboard: \textit{``[Shot 1, 0--3\,s] An extreme macro push-in under studio rim light: the 3-axis gimbal powers on and self-levels while the 2.0-inch OLED screen rotates 90\textdegree{} to landscape. Text: `1-inch CMOS sensor', `3-axis mechanical gimbal'. [Shot 2, 3--6.5\,s] A side tracking shot on a golden-hour esplanade: the creator's arm bounces while the gimbal keeps the camera level. Text: `Ultra-smooth cinematic stabilization', `4K\,/\,240\,fps'. [Shot 3, 6.5--10\,s] A slow orbit around the device on a stone pedestal against a twilight city. Text: `107\,GB built-in storage $|$ 2.0-inch OLED 1,000 nits'.''} (full storyboard in \cref{app:storyboard}).}

\label{fig:showcase}
\end{center}
\vspace{1em}
}]
{\renewcommand{\thefootnote}{\fnsymbol{footnote}}\footnotetext[1]{Equal contribution. \quad Contact: \href{mailto:kunyang.li@ucf.edu}{\texttt{kunyang.li@ucf.edu}}}}
\begin{abstract}
Recent advances in video generation have made automated advertising production increasingly feasible. However, an effective advertisement must faithfully represent the product and communicate its selling points through compelling visual storytelling, which requires more than visual synthesis: it calls for deeper product understanding and strategic decisions about how to present product value. Crucially, product research and creative design are closely intertwined, as the information gathered about a product determines which features can be faithfully demonstrated and how they should be presented. In this paper, we introduce \ours, the first end-to-end agentic framework for advertising video creation, transforming concise creative briefs into finished advertisements. Our multimodal agent autonomously researches products through web and image search and translates its findings into selling-point-driven storyboards that guide a frozen video generator.
To achieve this, we develop a dedicated data construction pipeline that yields two training datasets, AdProducer-SFT and AdProducer-RL, and train the agent through supervised fine-tuning followed by reinforcement learning with joint optimization of storyboard- and video-level rewards.
We further introduce AdCreation-Bench, comprising 305 advertising briefs across eight domains, with two complementary metrics: PG-Score for product correctness and creative execution, and Ad-Score for overall advertising effectiveness assessed by an independent judge. Experiments demonstrate that \ours improves the PG-Score of MiniMax H3 from 17.3 to 51.3 and generalizes to Veo~3.1 without additional training, improving its score from 28.0 to 35.9, with consistent gains in Ad-Score, VideoScore2, and human evaluations. Code, models, and data are fully released. 
\end{abstract}
    
\section{Introduction}
\label{sec:intro}

Recent advances in video generation have enabled the synthesis of highly realistic videos from text prompts and visual references~\cite{sora2024,wan2025,gao2025seedance,kling2025omni,liu2025phantom,jiang2025vace,li2026attendlocally,li2026packcache,alradi2026aeroworld,alradi2026keepsake,li2025gvd}, opening new opportunities for automated advertising production~\cite{iab2025videoP1,iab2025genai,xu2026nextads}. However, while conventional video generation primarily emphasizes visual quality and prompt alignment, advertising requires understanding what makes a product worth promoting and how to communicate its value through compelling visual storytelling~\cite{yang2026adspark,stewart1989executional,smith2007creativity}.

Consider creating an advertisement for a newly released foldable smartphone, as illustrated in \cref{fig:showcase}. A visually appealing video may still depict the wrong device, present incorrect specifications, or fail to showcase the folding experience that distinguishes the product. These failures highlight three key challenges in advertising video generation: preserving product identity, ensuring factual correctness, and translating selling points into meaningful visual demonstrations that convey product value~\cite{ftc1984substantiation,mcquarrie1999visual,teixeira2012emotion}.

Addressing these challenges requires more than powerful video generators; it calls for deeper product understanding and strategic decisions about how to present product value. Crucially, product research and creative design are closely intertwined, as the information gathered about a product determines which features can be faithfully demonstrated and how they should be presented. Recent advances in multimodal agents have demonstrated their potential for complex reasoning and tool use~\cite{wu2025mmsearchr1,feng2026gensearcher}, offering a promising direction for integrating these capabilities. This motivates us to explore whether a multimodal agent can learn to autonomously research products and design compelling advertising storyboards that guide a frozen video generator to produce advertisements from concise creative briefs.

\begin{figure}[t]
\centering
\includegraphics[width=\linewidth]{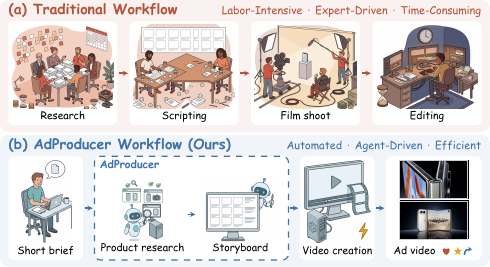}
\caption{\textbf{(a) Traditional advertising workflow vs.\ (b) \ours.} \ours automates product research and creative planning, enabling rapid advertising video creation from a short brief.}
\label{fig:teaser}
\vspace{-3pt}%
\end{figure}

\begin{figure*}[t]
\centering
\includegraphics[width=\textwidth]{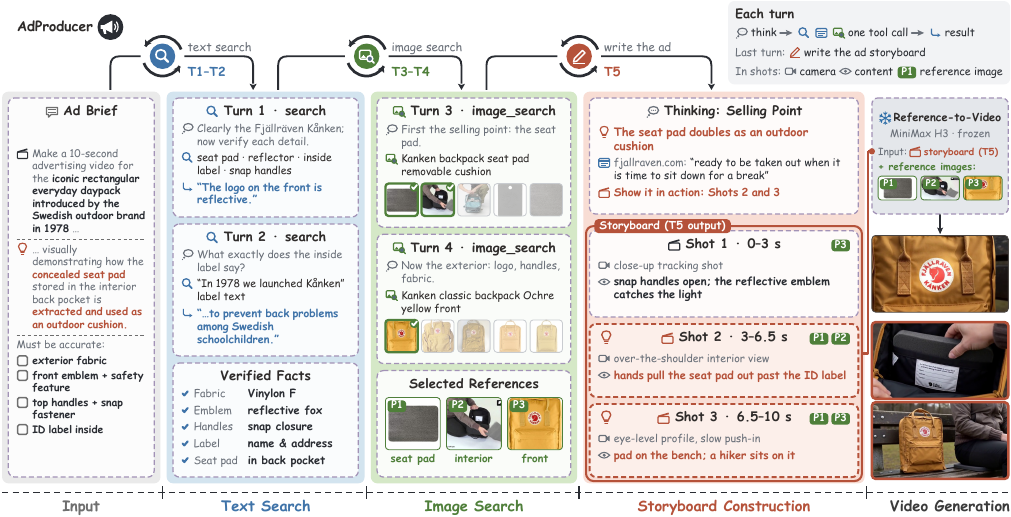}
\caption{\textbf{Inference trajectory of \ours on a benchmark brief.} The agent verifies product facts through text searches (T1--T2), selects visual references through image searches (T3--T4), and finally writes a selling-point-driven, three-shot storyboard (T5). Only the storyboard and reference images (P1--P3) are passed to a frozen reference-to-video generator. Prompt and storyboard details are in \cref{app:sb_kanken}.}
\label{fig:pipeline}
\end{figure*}

To this end, we introduce \textbf{\ours}, an end-to-end agentic framework that transforms concise briefs into advertising videos through autonomous product research and creative storyboard design (\cref{fig:teaser}b). To address the scarcity of training data, we develop a dedicated data construction pipeline. We first curate advertising briefs spanning 60 product categories across eight domains, then employ a teacher agent to conduct multi-round web and image searches. Grounded in the retrieved evidence and visual references, the teacher constructs selling-point-driven storyboards specifying shot content, timing, and camera movements. We subsequently filter the generated trajectories through structural and factual verification and render the storyboards into reference videos using MiniMax~H3~\cite{minimaxh3}. This pipeline produces two training datasets, \textbf{AdProducer-SFT} and \textbf{AdProducer-RL}, alongside \textbf{AdCreation-Bench}, a held-out benchmark comprising 305 advertising briefs. We further introduce two complementary evaluation metrics: \textbf{PG-Score} for product correctness and creative execution, and \textbf{Ad-Score} for overall advertising quality and effectiveness, assessed by an independent judge.

Building on these datasets, we adopt a two-stage training strategy consisting of supervised fine-tuning (SFT) followed by reinforcement learning (RL)~\cite{liu2026noisypref,liu2026agenticrorl}. SFT equips the agent with product research and creative storyboard design capabilities through teacher trajectory imitation, while RL further refines its decision-making through reward optimization. However, video-only rewards can provide noisy feedback due to the limitations of the underlying generator, whereas storyboard-only rewards cannot assess how effectively the planned content is realized. To address these limitations, we jointly optimize storyboard- and video-level rewards, providing complementary feedback on creative design and its visual realization.

Extensive experiments on AdCreation-Bench demonstrate that \ours substantially improves advertising video generation. With the video generator frozen and only the agent trained, \ours raises the PG-Score of MiniMax~H3 from 17.3 to 51.3, with strong gains in visual correctness, factual accuracy, and selling-point demonstration. Without retraining, the agent then transfers directly to Veo~3.1~\cite{veo3modelcard2025}, improving its PG-Score from 28.0 to 35.9. Evaluations using Ad-Score and VideoScore2~\cite{videoscore2}, together with human studies, further demonstrate improvements in advertising effectiveness and product fidelity, alongside competitive visual quality.

Our contributions are summarized as follows:
\begin{itemize}
    \item We introduce \ours, the first end-to-end agentic framework for advertising video creation from concise briefs, which integrates autonomous product research with selling-point-driven storyboard design.
    
    \item We develop two training datasets, AdProducer-SFT and AdProducer-RL, to address training data scarcity in agentic advertising video generation, and introduce AdCreation-Bench with PG-Score and Ad-Score to evaluate product correctness and advertising effectiveness.
    
    \item Experiments on AdCreation-Bench show that \ours substantially improves product correctness and selling-point presentation with MiniMax~H3 and further generalizes to Veo~3.1 without retraining. Ad-Score, VideoScore2, and human evaluations also confirm gains in advertising effectiveness and visual quality.
\end{itemize}
\section{Related Work}
\label{sec:related}

\paragraph{Advertising Generation.}
Prior work on advertising generation has primarily focused on images, exploring controllable layouts, product fidelity, and personalization~\cite{xu2026designyourad,shilova2023adbooster,gao2026creativeads,li2023planning,chen2025ctr}. Advertising video generation~\cite{xu2024anchorcrafter,bu2026agenticgen} introduces additional challenges in maintaining product consistency and factual accuracy across shots while effectively demonstrating selling points. Existing video-oriented studies have explored advertisement understanding and evaluation~\cite{hussain2017ads,long2025adsqa,zhang2025videoads,yang2026adspark}. However, end-to-end automated advertising video generation for real-world products remains relatively underexplored. In contrast, our work integrates autonomous product research with storyboard design to generate advertising videos directly from concise creative briefs.

\paragraph{Agentic Systems.}
LLM-based agents leverage multi-step reasoning and tool use to tackle complex, knowledge-intensive tasks~\cite{yao2023react,jin2025searchr1,chen2025research,zheng2025deepresearcher,li2025webthinker,wu2025mmsearchr1,wang2026adatooler,li2026cinemr,vayani2025alm}. Recent works have extended agentic approaches to image generation through knowledge retrieval and iterative refinement~\cite{jiang2026genagent,zhao2026toolartist,feng2026gensearcher}, and to multi-shot video generation through script planning, storyboarding, and scene-level coordination~\cite{lin2023videodirectorgpt,xie2024dreamfactory,yuan2024mora,wu2025movieagent,xu2025filmagent,li2026videogenagent,song2026codirector,wang2026videocoder}. In advertising, existing agentic approaches primarily explore reward-guided video generation using business feedback and quality signals~\cite{bu2026agenticgen}. In contrast, our work focuses on autonomous product research and selling-point-driven storyboard design, enabling advertising video generation from concise briefs.

\vspace{3.8pt}%
\section{Method}
\label{sec:method}

\subsection{Overview}
\label{sec:framework}

Given a concise creative brief $b$, \ours aims to enable end-to-end advertising video generation, automatically transforming a user's request into a finished advertisement without manual intervention. To achieve this, we introduce a multimodal agent $\pi_\theta$ that autonomously researches the target product, gathers visual references, and develops a structured multi-shot storyboard with selected reference images. A frozen video generator $\mathcal G$ then renders the storyboard into the final advertisement $v$. The agent's interleaved reasoning and tool use form a trajectory $\tau\sim\pi_\theta(\cdot\mid b)$, which culminates in the storyboard and its selected reference images.

The agent iteratively reasons over gathered evidence to determine its next actions, using three tools: \texttt{search} to identify products and verify facts, \texttt{browse} to examine relevant webpages with Gemini~3.1 Pro~\cite{gemini31pro}, and \texttt{image\_search} to retrieve product-specific visual references. Based on the collected evidence, the agent designs a structured storyboard (\cref{fig:pipeline}) specifying shot timing, visual content, camera movements, and reference assignments. This evidence-grounded design preserves product identity across shots, ensures accurate on-screen text, and shows selling points through visual actions.

\vspace{5pt}%
\subsection{Training}
\label{sec:train}

\noindent\textbf{Two-Stage Training.}\label{sec:sft}
We initialize \ours from Qwen3.5-9B~\cite{qwen35} and adopt a two-stage training strategy. We first perform supervised fine-tuning on AdProducer-SFT, teaching the agent to conduct multi-round product research, select visual references, and design selling-point-driven storyboards by imitating teacher trajectories. We then further optimize the agent on AdProducer-RL through reinforcement learning, encouraging more effective tool use and creative decision-making beyond teacher imitation. Throughout training, the video generator $\mathcal{G}$ remains frozen, and only the multimodal agent is optimized.

\noindent\textbf{Joint Reward Design.}\label{sec:rl}
Directly evaluating generated advertisements provides feedback on the final outcome, but can be unreliable, as video quality depends not only on the agent's decisions but also on the limitations and stochasticity of the frozen video generator. Conversely, evaluating storyboards alone cannot determine whether the planned content is faithfully realized in the final video. To address these complementary limitations, we introduce a joint reward that considers both video and storyboard quality.

We first render each storyboard using the frozen generator $\mathcal{G}$ and compute the video reward $R_{\mathrm{video}}=\mathrm{PG}(v)$ (\cref{eq:video}), assessing how effectively the planned content is realized in the final advertisement.

To complement this outcome-based feedback, we employ Gemini~3.1 Pro to evaluate the storyboard and selected references against the brief, verified product facts, official images, and reference video. The judge assesses product identity, factual correctness, required on-screen text, and selling-point demonstration, while considering shot feasibility and narrative coherence. The storyboard reward $R_{\mathrm{sb}}\in\{0,0.25,0.5,0.75,1\}$ applies strict score caps: incorrect product identity or contradictory facts limit the reward to $0.25$, while missing required text or inadequate selling-point demonstration limit it to $0.5$. This provides direct feedback on the agent's research and creative storyboard design, independent of rendering errors.
The two rewards are then combined into the joint reward:
\begingroup\setlength{\abovedisplayskip}{8pt}\setlength{\belowdisplayskip}{8pt}%
\begin{equation}
R=(1-\alpha)\,R_{\mathrm{video}}+\alpha\,R_{\mathrm{sb}},
\label{eq:reward}
\end{equation}
\endgroup
where $\alpha=0.5$ balances the two reward components.

\noindent\textbf{Optimization.}
We optimize the agent using Group Relative Policy Optimization (GRPO)~\cite{shao2024deepseekmath}. For each brief $b$, we sample a group of $K$ trajectories $\{\tau_i\}_{i=1}^{K}$ from the old policy $\pi_{\theta_{\mathrm{old}}}$ with live tool execution, and normalize each reward within the group to obtain the advantage:
\begin{equation}
\hat{A}_i=
\frac{R_i-\operatorname{mean}(\{R_j\}_{j=1}^{K})}
{\operatorname{std}(\{R_j\}_{j=1}^{K})}.
\label{eq:adv}
\end{equation}
The policy is then updated by maximizing
\begin{equation}
\begin{aligned}
\mathcal{J}(\theta)=\mathbb{E}_{b,\{\tau_i\}}\Bigg[
\frac{1}{K}\sum_{i=1}^{K}
\frac{1}{|\mathcal{T}_i|}\sum_{t\in\mathcal{T}_i}
\min\Big(
r_{i,t}(\theta)\hat{A}_i,\\
\operatorname{clip}\big(r_{i,t}(\theta),
1-\varepsilon_{\mathrm{low}},
1+\varepsilon_{\mathrm{high}}\big)\hat{A}_i
\Big)
\Bigg],
\end{aligned}
\label{eq:grpo}
\end{equation}
where $\mathcal{T}_i$ denotes agent-generated tokens, and we use asymmetric clipping with $\varepsilon_{\mathrm{low}}=0.2$ and $\varepsilon_{\mathrm{high}}=0.28$~\cite{yu2025dapo}.

\section{Training Data Construction}
\label{sec:data}

\begin{figure}[!t]
\centering
\includegraphics[width=\linewidth]{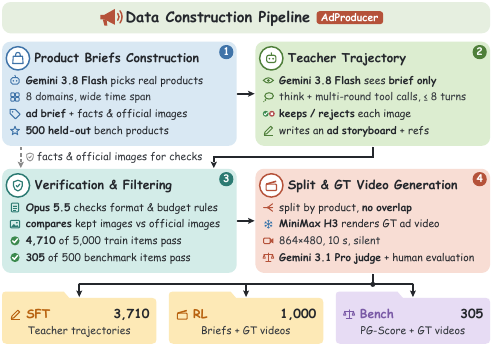}
\caption{\textbf{Overview of the four-stage data construction pipeline} for training and evaluation of agentic advertising video generation.}
\label{fig:datapipe}
\end{figure}

\begin{figure*}[t]
\centering
\includegraphics[width=\textwidth]{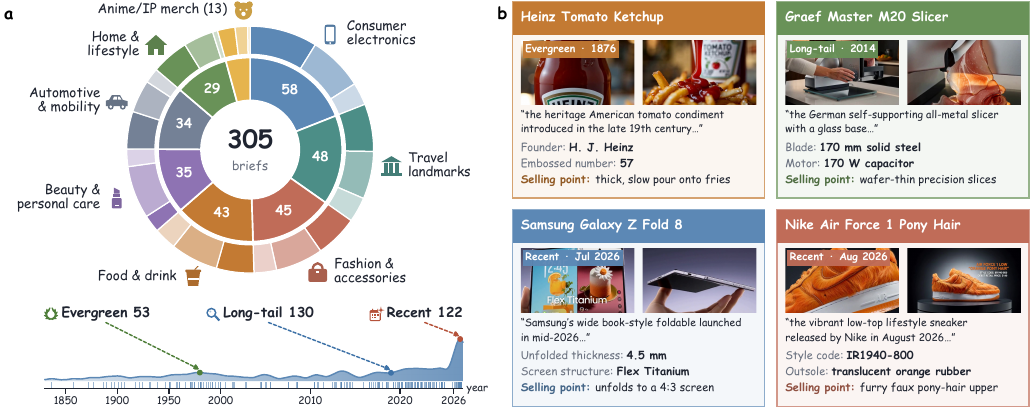}
\caption{\textbf{AdCreation-Bench overview.}
(a)~305 briefs over eight domains, covering a broad time span from 1836 to 2026.
(b)~Example items across different categories and release times, each with a brief excerpt, key product facts and the selling point to demonstrate.}
\label{fig:bench}
\end{figure*}

Training an agent for end-to-end advertising generation requires data that capture the complete workflow of product research, visual reference selection, and storyboard design. However, existing datasets rarely provide such multi-step trajectories. To address this gap, we develop a four-stage data construction pipeline (\cref{fig:datapipe}) that produces high-quality training data through multi-round information retrieval, evidence verification, and storyboard construction.

\noindent\textbf{\ding{172} Product Briefs Construction.}
We select products from eight advertising domains spanning 60 categories: consumer electronics, fashion and accessories, food and beverage, beauty and personal care, home and lifestyle, automotive and mobility, travel destinations, and anime and IP merchandise. These domains require diverse visual presentations, from demonstrating product functionality to highlighting material textures and showcasing destinations.

For each product, we use Gemini 3.8 Flash~\cite{gemini38flash}, equipped with web and image search, to construct a creative brief specifying the product, a key selling point suitable for a 10-second advertisement, and factual requirements without revealing their values. We additionally collect a source-backed fact table, distinguishing attributes for disambiguating similar products, and two to five official reference images. These supporting materials remain hidden from the agent and are used exclusively for verification, filtering (\ding{174} and \ding{175}), reward computation, and evaluation.

\noindent\textbf{\ding{173} Teacher Trajectory Generation.}
We employ Gemini 3.8 Flash~\cite{gemini38flash} as the teacher agent, which receives only the constructed brief and system prompt and operates with the same tools and budget as the student agent (\cref{sec:framework}). During multi-round web and image searches, the teacher progressively gathers product evidence, evaluates the relevance of retrieved images, and determines how to translate product features into compelling visual presentations. Each retrieved image is retained or discarded with a justification and, when selected, assigned to a corresponding shot.

Once the research is complete, the teacher constructs the storyboard in the final turn, grounded in the retrieved evidence. The storyboard binds each recurring subject to its reference images and specifies a timed sequence of at least three shots over the 10-second ad, each with its composition, action, camera movement, referenced images, and on-screen text. Its design follows advertising conventions: the first shot captures attention, the middle shots visibly demonstrate the selling point rather than merely showing the product, and the final shot leaves a memorable product impression, with all on-screen text quoting verified facts (examples in \cref{fig:pipeline} and \cref{app:storyboard}). These trajectories preserve the reasoning, tool calls, and observations behind each storyboard.

\noindent\textbf{\ding{174} Trajectory Verification and Filtering.}
To ensure the quality and reliability of the collected trajectories, we adopt a two-stage filtering process, consisting of an initial validation during data collection and a subsequent verification using Claude Opus 5.5~\cite{claudeopus55}. Specifically, we combine rule-based checks with model-based verification to identify invalid tool calls, missing or mismatched visual references, and storyboards that fail to meet the required output format. We further assess product consistency by comparing selected reference images against official product images, filtering out trajectories that retrieve incorrect products or variants. Through this process, we retain high-quality trajectories that satisfy structural requirements and maintain factual and visual consistency with the target products.

\noindent\textbf{\ding{175} Reference Video Generation.}
To provide visual references for reinforcement learning and benchmark evaluation, we employ the frozen MiniMax H3 video generator~\cite{minimaxh3} to synthesize advertising videos from the verified storyboards. Since the generated videos may contain visual or factual inaccuracies, we assess product fidelity and factual correctness against official product images and annotated facts, rather than relying solely on the synthesized videos.

After verification and filtering, we retain 4{,}710 valid samples from 5{,}000 training candidates, comprising 3{,}710 trajectories for supervised fine-tuning (AdProducer-SFT) and 1{,}000 examples for reinforcement learning (AdProducer-RL). We additionally curate AdCreation-Bench, comprising 305 briefs selected from 500 candidates, with no product overlap between training and evaluation sets (details in \cref{sec:bench}).

\vspace{-4pt}%
\section{AdCreation-Bench}
\label{sec:bench}

Existing video generation benchmarks primarily focus on visual fidelity and temporal consistency~\cite{vbench,vbench2,vbenchpp}. However, a visually appealing video does not necessarily make an effective advertisement: it must also faithfully represent the advertised product, convey accurate information, and communicate its selling points through compelling visual storytelling. To evaluate these requirements, we introduce \textbf{AdCreation-Bench}, which consists of creative briefs with verified product annotations (\cref{sec:bench_construct}) and two complementary metrics, PG-Score for product correctness and Ad-Score for advertising effectiveness (\cref{sec:bench_metrics}).

\subsection{Benchmark Construction}
\label{sec:bench_construct}
Following the pipeline in \cref{sec:data}, we curate 305 evaluation samples, each featuring a unique product with no overlap with training data. To assess different product knowledge challenges, we organize the benchmark~\cref{fig:bench}  into three groups: \textit{recent} (122), comprising products released between July and October 2026, after the models' knowledge cutoffs; \textit{long-tail} (130), covering less-prominent or sparsely documented products released from 1903 to 2026, including older products reassigned from the recent group after launch-date verification; and \textit{evergreen} (53), focusing on lesser-known component-level details or secondary specifications of widely recognized products released from 1876 to 2022. Each brief requires a median of six factual details to be retrieved, and 92\% of samples require accurate on-screen text. For briefs without required on-screen text, the text accuracy term is excluded and the remaining weights are renormalized. All reference videos undergo verification by Gemini 3.1 Pro~\cite{gemini31pro} and human inspection.

\subsection{Evaluation Metrics}
\label{sec:bench_metrics}

\noindent\textbf{PG-Score.}
We introduce the \textbf{Product-Grounding Score} to evaluate generated advertising videos rather than intermediate storyboards. We employ Gemini~3.1 Pro~\cite{gemini31pro} as a multimodal judge to assess the generated videos against the brief, hidden factual annotations, official product images, and a reference video that serves as an aesthetic reference.

PG-Score comprises five dimensions: \textit{visual correctness} (vis), measuring product identity and appearance; \textit{text accuracy} (txt), assessing factual correctness of on-screen text; \textit{selling point} (sell), measuring visual demonstration of product benefits; \textit{shot design} (shot), assessing composition and coherence; and \textit{aesthetics} (aes), evaluating overall visual quality. Each dimension is scored on a discrete scale within $[0,1]$, and the overall score is computed as:
\begingroup\setlength{\abovedisplayskip}{5.6pt}\setlength{\belowdisplayskip}{5.6pt}%
\begin{equation}
\text{PG}(v)=0.4q_{\text{vis}}+0.2q_{\text{txt}}+0.2q_{\text{sell}}+0.1q_{\text{shot}}+0.1q_{\text{aes}}.
\label{eq:video}
\end{equation}
\endgroup

\begin{table*}[t]
\centering
\small
\setlength{\tabcolsep}{2.9pt}
\caption{\textbf{PG-Score comparison of \ours with proprietary and open-source video generators on AdCreation-Bench.} \ours consistently improves advertising video quality across different video generators, demonstrating its effectiveness and generalizability.}
\label{tab:pg_main}
\begin{tabular}{@{}llccccccccc@{}}
\toprule
& & \multicolumn{5}{c}{PG-Score dimensions $\uparrow$} & \multicolumn{3}{c}{PG-Score by regime $\uparrow$} & \\
\cmidrule(lr){3-7}\cmidrule(lr){8-10}
& Method & \shortstack{Visual\\correctness} & \shortstack{Text\\accuracy} & \shortstack{Selling\\point} & \shortstack{Shot\\design} & \raisebox{0.5\normalbaselineskip}{Aesthetics} & \raisebox{0.5\normalbaselineskip}{Recent} & \raisebox{0.5\normalbaselineskip}{Long-tail} & \raisebox{0.5\normalbaselineskip}{Evergreen} & \shortstack{\textbf{PG-Score}\\\textbf{Overall} $\uparrow$}\\
\midrule
\multirow{3}{*}{Proprietary}
& Veo~3.1 Lite~\cite{veo3modelcard2025} & 5.2 & 15.3 & \underline{58.4} & 53.7 & 54.6 & 24.9 & 28.7 & 30.8 & 27.8\\
& Veo~3.1 Fast~\cite{veo3modelcard2025} & 5.1 & \underline{16.9} & 56.2 & \underline{57.9} & 57.9 & 26.5 & 29.8 & 28.6 & 28.4\\
& Veo~3.1~\cite{veo3modelcard2025} & 6.3 & 15.1 & 56.1 & 55.6 & 55.6 & 24.2 & 29.5 & 31.3 & 28.0\\
\midrule
\multirow{6}{*}{Open}
& CogVideoX1.5-5B~\cite{yang2025cogvideox} & 0.7 & 0.0 & 4.5 & 2.0 & 1.0 & 0.8 & 2.5 & 0.8 & 1.5\\
& HunyuanVideo-13B~\cite{kong2024hunyuanvideo} & 0.8 & 0.0 & 3.2 & 4.4 & 35.2 & 4.5 & 5.0 & 6.6 & 5.1\\
& LTX-2-19B~\cite{hacohen2026ltx2} & 0.5 & 0.0 & 6.5 & 7.2 & 16.9 & 3.2 & 4.8 & 4.0 & 4.0\\
& Mochi-1-10B~\cite{genmo2024mochi} & 1.2 & 0.0 & 8.8 & 5.2 & 12.1 & 3.2 & 4.3 & 5.5 & 4.1\\
& Wan2.2-T2V-A14B~\cite{wan2025} & 1.8 & 0.1 & 12.7 & 6.6 & 36.6 & 6.6 & 8.4 & 9.0 & 7.8\\
& MiniMax H3~\cite{minimaxh3} & 3.5 & 1.3 & 34.1 & 39.7 & 44.9 & 16.7 & 17.5 & 18.3 & 17.3\\
\midrule
\oursrow & \textbf{\ours + Veo~3.1} & \underline{25.7} & 15.2 & 55.4 & 50.0 & \underline{61.2} & \underline{34.6} & \underline{36.5} & \underline{37.4} & \underline{35.9}\\
\oursrow \multirow{-2}{*}{\textbf{Ours}} & \textbf{\ours + MiniMax H3} & \textbf{36.5} & \textbf{40.4} & \textbf{71.6} & \textbf{64.9} & \textbf{76.2} & \textbf{56.8} & \textbf{46.7} & \textbf{49.6} & \textbf{51.3}\\
\bottomrule
\end{tabular}
\end{table*}

\noindent\textbf{Ad-Score.}
PG-Score evaluates a video mainly from the product side, \ie, whether the product is correctly presented and its selling points are demonstrated. A good advertisement, however, should also be judged as a whole from the viewer's perspective: whether it captures attention, tells a coherent story, and persuades the audience. We therefore introduce the \textbf{Advertising Quality Score (Ad-Score)}, which assesses overall advertising quality from a viewer-oriented perspective. Its dimensions are distilled from the criteria by which advertisements are commonly evaluated in the advertising industry and research~\cite{teixeira2012emotion,smith2007creativity,stewart1989executional}, and are organized into three groups: \textit{brief fidelity} and \textit{product presentation} (brief and product); \textit{visual quality} and \textit{narrative flow} (craft); and \textit{hook}, \textit{creative quality}, \textit{persuasiveness}, and \textit{purchase intent} (advertising effectiveness).

Ad-Score is computed by a separate judge, GPT-6 Sol~\cite{gpt6sol}, which is not involved in training and has no access to the reference video. Given the generated video, creative brief, verified product facts, and official images, the judge rates each dimension on an anchored 1--5 scale and assigns a holistic 1--10 score, which we report as the Ad-Score instead of averaging the dimension scores. Together, PG-Score and Ad-Score evaluate whether an advertisement is both faithful to the product and effective for its viewers.

\section{Experiments}
\label{sec:exp}

\subsection{Experimental Setup}
\label{sec:setup}

\begin{table}[t]
\centering
\small
\setlength{\tabcolsep}{0.5pt}
\caption{\textbf{Human evaluation of advertising video quality.} \ours achieves the highest ratings across all dimensions, with substantial gains in product fidelity and informativeness.}

\label{tab:human}
\begin{tabular}{@{}lccccc@{}}
\toprule
Method & \shortstack{Visual\\quality} & \shortstack{Product\\fidelity} & \shortstack{Informa-\\tiveness} & \shortstack{Ad\\attitude} & \shortstack{Purchase\\intention}\\
\midrule
Veo~3.1 & 3.92$_{\pm.26}$ & 2.88$_{\pm.46}$ & 3.50$_{\pm.46}$ & 3.49$_{\pm.32}$ & 3.29$_{\pm.42}$\\
HunyuanVideo-13B & 2.21$_{\pm.31}$ & 1.25$_{\pm.18}$ & 1.29$_{\pm.22}$ & 1.50$_{\pm.26}$ & 1.21$_{\pm.17}$\\
Wan2.2-T2V-A14B & 2.21$_{\pm.31}$ & 1.42$_{\pm.20}$ & 1.38$_{\pm.26}$ & 1.72$_{\pm.25}$ & 1.38$_{\pm.20}$\\
MiniMax H3 & 3.46$_{\pm.31}$ & 2.46$_{\pm.39}$ & 2.50$_{\pm.46}$ & 2.53$_{\pm.36}$ & 2.58$_{\pm.41}$\\
\midrule
\oursrow \textbf{\ours + H3} & \textbf{4.21}$_{\pm.26}$ & \textbf{4.75}$_{\pm.18}$ & \textbf{4.67}$_{\pm.23}$ & \textbf{4.07}$_{\pm.26}$ & \textbf{3.96}$_{\pm.28}$\\
\bottomrule
\end{tabular}
\end{table}

\begin{table}[t]
\centering
\small
\setlength{\tabcolsep}{4pt}
\caption{\textbf{VideoScore2 of general video quality.} \ours maintains competitive visual quality while improving text-video alignment and physical plausibility across frozen video generators.}

\label{tab:videoscore2}
\begin{tabular}{@{}lccc@{}}
\toprule
Method & \shortstack{Visual\\quality} & \shortstack{Text\\alignment} & \shortstack{Physical\\sense}\\
\midrule
Veo~3.1~\cite{veo3modelcard2025} & \underline{4.66} & 4.01 & \underline{4.33}\\
\midrule
CogVideoX1.5-5B~\cite{yang2025cogvideox} & 3.10 & 2.30 & 2.82\\
HunyuanVideo-13B~\cite{kong2024hunyuanvideo} & 3.97 & 2.38 & 3.60\\
Wan2.2-T2V-A14B~\cite{wan2025} & 4.16 & 2.78 & 3.80\\
MiniMax H3~\cite{minimaxh3} & 4.24 & 3.53 & 3.96\\
\midrule
\oursrow \textbf{\ours + Veo~3.1} & \textbf{4.67} & \underline{4.03} & \textbf{4.35}\\
\oursrow \textbf{\ours + MiniMax H3} & 4.32 & \textbf{4.09} & 4.11\\
\bottomrule
\end{tabular}
\end{table}

\paragraph{Implementation Details.}
We train \ours with SFT (learning rate $1\times10^{-5}$, batch size 16) on 4 NVIDIA H100 GPUs. RL is conducted on 4 H100 GPUs with a learning rate of $1\times10^{-6}$, sampling $K=6$ trajectories per brief with 8 briefs per step, while another 4 H100 GPUs serve MiniMax~H3 for video generation. Each trajectory is limited to 8 tool calls (at most 2 image searches, each returning up to 5 images), 6K tokens per turn, and a 64K-token context; trajectories exceeding these limits are truncated. The \texttt{browse} tool is powered by Gemini~3.1 Pro. At inference, we use temperature 0.6 and top-$p$ 0.9. Veo~3.1 generates 720p videos, while other generators produce 10-second 480p videos, all in 16:9 without audio.

\paragraph{Baselines.}
We compare \ours with state-of-the-art video generators that are directly prompted with the brief: the proprietary Veo~3.1 family (Lite, Fast, and standard)~\cite{veo3modelcard2025}, and the open-source CogVideoX1.5-5B~\cite{yang2025cogvideox}, HunyuanVideo-13B~\cite{kong2024hunyuanvideo}, LTX-2-19B~\cite{hacohen2026ltx2}, Mochi-1-10B~\cite{genmo2024mochi}, Wan2.2-T2V-A14B~\cite{wan2025}, and MiniMax~H3~\cite{minimaxh3}. The baselines generate the video directly from the brief, whereas \ours first interprets the brief, researches the product, and writes a storyboard with reference images, which a frozen generator (MiniMax~H3 or Veo~3.1) then renders. Overall, all methods follow the same protocol for a fair comparison: each takes a creative brief as input and outputs an advertising video for evaluation.

\paragraph{Evaluation.}
We evaluate all methods with the two metrics of AdCreation-Bench (\cref{sec:bench_metrics}): PG-Score, which assesses product correctness and selling-point demonstration against verified annotations, and Ad-Score, which assesses overall advertising quality from a viewer-oriented perspective. If \ours's output fails the strict format check, we fall back to a lenient check and, if that also fails, to the baseline brief-to-video generation.

We further conduct a human study following the absolute category rating (ACR) protocol of ITU-T P.910~\cite{itu_p910_2023}, in which participants rate the videos of five methods on eight briefs from five domains (40 clips). Each clip is shown alone with its brief and an official product image; method names are hidden, the order is randomized per participant, and repeated clips serve as attention checks. The questionnaire adopts established measures from advertising research and rates five dimensions on labelled five-point scales: \textit{visual quality} (the ACR scale), \textit{product fidelity}, \textit{informativeness}, \ie, whether the ad conveys the product's key features~\cite{ducoffe1996value}, \textit{attitude toward the ad}~\cite{mackenzie1986aad,mitchell1981aad}, and \textit{purchase intention}~\cite{spears2004measuring}. Together, these dimensions cover both whether the product is shown correctly and how the advertisement is perceived by viewers. We report mean opinion scores and measure inter-rater agreement with Krippendorff's $\alpha$~\cite{hayes2007alpha}.

Finally, we report VideoScore2~\cite{videoscore2} as a complementary check of visual quality, text-video alignment, and physical plausibility, verifying that optimizing for advertising does not degrade video quality.

\vspace{6pt}%
\subsection{Main Results on AdCreation-Bench}
\label{sec:results}

\begin{table*}[t]
\centering
\small
\setlength{\tabcolsep}{3.6pt}
\caption{\textbf{Ad-Score comparison on AdCreation-Bench.} \ours consistently improves overall ad quality across different frozen video generators, demonstrating stronger product presentation and advertising effectiveness.}
\label{tab:adjudge}
\begin{tabular}{@{}llccccccccc@{}}
\toprule
& & \multicolumn{2}{c}{Brief \& product $\uparrow$} & \multicolumn{2}{c}{Craft $\uparrow$} & \multicolumn{4}{c}{Ad effectiveness $\uparrow$} & \\
\cmidrule(lr){3-4}\cmidrule(lr){5-6}\cmidrule(lr){7-10}
& Method & \shortstack{Brief\\fidelity} & \shortstack{Product\\presentation} & \shortstack{Visual\\quality} & \shortstack{Narrative\\flow} & Hook & \shortstack{Creative\\quality} & \shortstack{Persua-\\siveness} & \shortstack{Purchase\\intent} & \shortstack{\textbf{Ad-Score}\\\textbf{(1--10)} $\uparrow$}\\
\midrule
\multirow{3}{*}{Proprietary}
 & Veo~3.1 Lite~\cite{veo3modelcard2025} & 2.09 & 1.41 & 2.27 & 2.98 & 3.43 & 2.78 & 2.87 & 1.93 & 2.79\\
 & Veo~3.1 Fast~\cite{veo3modelcard2025} & 2.15 & 1.43 & 2.29 & 3.02 & \underline{3.50} & 2.77 & 2.85 & 1.94 & 2.87\\
 & Veo~3.1~\cite{veo3modelcard2025} & 2.16 & 1.45 & 2.33 & \underline{3.03} & \textbf{3.52} & 2.81 & 2.85 & 1.93 & 2.87\\
\midrule
\multirow{6}{*}{Open}
 & CogVideoX1.5-5B~\cite{yang2025cogvideox} & 1.97 & 1.11 & 1.60 & 1.51 & 1.85 & 1.54 & 1.41 & 1.34 & 1.39\\
 & HunyuanVideo-13B~\cite{kong2024hunyuanvideo} & 1.91 & 1.06 & \underline{3.04} & 1.46 & 2.18 & 1.82 & 1.65 & 1.54 & 1.60\\
 & LTX-2-19B~\cite{hacohen2026ltx2} & 1.95 & 1.08 & 2.24 & 1.53 & 2.08 & 1.70 & 1.62 & 1.42 & 1.50\\
 & Mochi-1-10B~\cite{genmo2024mochi} & 2.03 & 1.15 & 2.41 & 1.36 & 2.30 & 1.91 & 1.88 & 1.59 & 1.75\\
 & Wan2.2-T2V-A14B~\cite{wan2025} & 2.04 & 1.20 & 2.77 & 1.23 & 2.28 & 1.89 & 2.02 & 1.63 & 1.73\\
 & MiniMax H3~\cite{minimaxh3} & 2.12 & 1.35 & 2.53 & 2.61 & 2.81 & 2.55 & 2.60 & 1.90 & 2.72\\
\midrule
\oursrow & \textbf{\ours + Veo~3.1} & \underline{2.40} & \underline{2.15} & 2.39 & 2.94 & 3.27 & \underline{2.85} & \underline{2.92} & \underline{2.19} & \underline{3.47}\\
\oursrow \multirow{-2}{*}{\textbf{Ours}} & \textbf{\ours + MiniMax H3} & \textbf{2.66} & \textbf{2.94} & \textbf{3.45} & \textbf{3.36} & 3.46 & \textbf{3.03} & \textbf{3.33} & \textbf{2.66} & \textbf{4.70}\\
\bottomrule
\end{tabular}
\end{table*}

\begin{table}[t]
\centering
\small
\setlength{\tabcolsep}{3.3pt}
\caption{\textbf{Comparison of untrained prompting workflows and task-specific agent training.} MiniMax H3 is as the video generator. \ours outperforms larger untrained agents under both planning-only and search-enabled workflows.}

\label{tab:workflow}
\begin{tabular}{@{}lcccccc@{}}
\toprule
Agent / Model & \shortstack{Vis.\\corr.} & \shortstack{Text\\acc.} & \shortstack{Sell.\\point} & \shortstack{Shot\\design} & \shortstack{Aes-\\thet.} & \shortstack{\textbf{PG-}\\\textbf{Score}}\\
\midrule
\oursrow \multicolumn{7}{@{}l}{\textit{Untrained agents $\cdot$ Planning-only (no retrieval)}}\\
\hspace{0.6em}Qwen3.5-9B & 1.7 & 3.3 & 40.0 & 40.0 & 46.7 & 18.0\\
\hspace{0.6em}Qwen3.5-35B-A3B & 1.7 & 8.3 & 45.8 & 46.7 & 48.3 & 21.0\\
\hspace{0.6em}Qwen3-30B-A3B & 3.3 & 10.8 & 56.7 & 61.7 & 58.3 & 26.8\\
\oursrow \multicolumn{7}{@{}l}{\textit{Untrained agents $\cdot$ Search + Planning}}\\
\hspace{0.6em}Qwen3.5-9B & 11.7 & 10.8 & 40.8 & 40.0 & 48.3 & 23.8\\
\hspace{0.6em}Qwen3.5-35B-A3B & 30.0 & 17.5 & 45.0 & 50.0 & 51.7 & 34.7\\
\hspace{0.6em}Qwen3-VL-30B-A3B & 0.8 & 2.5 & 29.2 & 30.0 & 36.7 & 13.3\\
\oursrow \multicolumn{7}{@{}l}{\textit{Task-specific training}}\\
\oursrow \hspace{0.6em} \textbf{\ours-9B} & \textbf{36.5} & \textbf{40.4} & \textbf{71.6} & \textbf{64.9} & \textbf{76.2} & \textbf{51.3}\\
\bottomrule
\end{tabular}
\vspace{-3pt}%
\end{table}

\begin{figure*}[t]
\centering
\includegraphics[width=\textwidth]{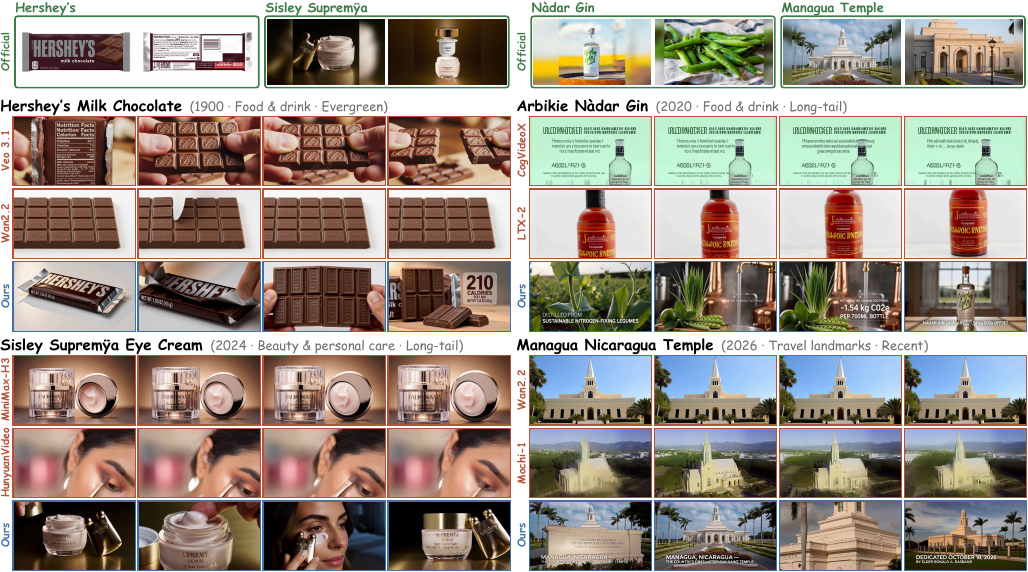}
\caption{\textbf{Qualitative comparison across video generators and product categories.} Red: direct-prompting baselines; blue: AdProducer; green: official product images. Grounded in product research, AdProducer depicts each product as it really looks, presents verified facts as legible on-screen text, and unfolds the selling point across multiple shots with varied compositions. Detailed analysis in \cref{sec:ablation}.}

\label{fig:qual}
\end{figure*}

\paragraph{PG-Score.}
Directly prompted generators struggle with product correctness, scoring at most 6.3 in visual correctness and 16.9 in text accuracy (\cref{tab:pg_main}). With frozen MiniMax H3, \ours improves PG-Score from 17.3 to 51.3, with the largest gains in visual correctness, text accuracy, and selling-point presentation, particularly for recent products, highlighting the value of external product evidence and visual references. \ours also generalizes to Veo~3.1, improving PG-Score from 28.0 to 35.9. The smaller gain reflects Veo~3.1's weaker adherence to reference images, suggesting that the final quality also depends on how faithfully the frozen generator follows the storyboard and references. Qualitative examples in \cref{fig:qual} further show more faithful products and clearer selling points.

\paragraph{Ad-Score.}
As shown in \cref{tab:adjudge}, \ours + H3 achieves the highest overall Ad-Score of 4.70, compared with 2.87 for the strongest baseline. With the same frozen generator, \ours raises the Ad-Score of MiniMax~H3 from 2.72 to 4.70. The largest gain in product presentation highlights the value of product-specific evidence, while improvements in narrative flow, persuasiveness, and purchase intent suggest that storyboard design also enhances advertising appeal. However, hook quality remains slightly below Veo~3.1, suggesting that capturing immediate viewer attention remains challenging. \ours also improves Veo~3.1's overall score from 2.87 to 3.47, demonstrating benefits across different generators.

\subsection{Complementary Results}
\label{sec:complementary}

\paragraph{Human Evaluation.}
We collect questionnaires from 15 participants. As shown in \cref{tab:human}, \ours + H3 achieves the highest ratings across all dimensions, with particularly large gains in product fidelity (4.75) and informativeness (4.67), highlighting the benefits of evidence-grounded product research and storyboard design. Improvements in ad attitude (4.07) and purchase intention (3.96) further indicate that these gains extend to viewers' perceptions of advertising appeal and persuasiveness. Krippendorff's $\alpha$ of 0.68--0.88 indicates substantial inter-rater agreement.

\begin{table}[t]
\centering
\small
\setlength{\tabcolsep}{0.8pt}
\caption{\textbf{Joint reward ablation} with MiniMax H3 as video generator. Combining storyboard and video rewards yields the largest improvement over SFT, outperforming single-reward variant.}

\label{tab:ablation}
\begin{tabular}{@{}lcccccc@{}}
\toprule
Method & \shortstack{Vis.\\corr.} & \shortstack{Text\\acc.} & \shortstack{Sell.\\point} & \shortstack{Shot\\design} & \shortstack{Aes-\\thet.} & \shortstack{\textbf{PG-}\\\textbf{Score}}\\
\midrule
MiniMax H3 & 3.5 & 1.3 & 34.1 & 39.7 & 44.9 & 17.3\\
\hspace{0.6em}+ \ours-SFT & 34.5 & 35.3 & 68.7 & 64.3 & 72.3 & 48.4\\
\hspace{0.6em}+ \ours-RL w.o.\ $R_{\mathrm{video}}$ & 35.4 & 39.9 & 68.8 & 63.5 & 74.8 & 49.7\\
\hspace{0.6em}+ \ours-RL w.o.\ $R_{\mathrm{sb}}$ & 35.3 & 39.0 & 67.6 & 64.6 & 76.1 & 49.5\\
\oursrow \hspace{0.6em}+ \ours-RL & \textbf{36.5} & \textbf{40.4} & \textbf{71.6} & \textbf{64.9} & \textbf{76.2} & \textbf{51.3}\\
\bottomrule
\end{tabular}
\end{table}

\paragraph{VideoScore2.}
VideoScore2 results (\cref{tab:videoscore2}) show that \ours improves general video quality without modifying the underlying generators. With MiniMax~H3, text-video alignment increases from 3.53 to 4.09, alongside gains in visual quality (4.24 to 4.32) and physical plausibility (3.96 to 4.11). These gains can be attributed to better-grounded references and more coherent shot-level instructions. %

\subsection{Ablation Studies and Analysis}
\label{sec:ablation}

\paragraph{Agentic Workflow Analysis.}
We investigate whether prompting and larger pretrained models can replace task-specific agent training (\cref{tab:workflow}). We compare untrained Qwen agents under two workflows: planning-only, which explicitly guides selling-point reasoning and storyboard construction without external tools, and search-enabled, which additionally enables product research and visual reference retrieval. Planning-only agents achieve reasonable selling-point and shot-design scores but struggle with visual correctness. Search further improves product grounding (\eg, 34.7 for Qwen3.5-35B-A3B). In contrast, our trained 9B agent achieves 51.3, substantially outperforming both its untrained counterpart (+27.5) and larger untrained models (+16.6 over the best). These results highlight the importance of task-specific training beyond carefully designed prompts, external tool access, and model scaling.

\paragraph{Joint Reward Ablation.}
We examine the contributions of storyboard and video rewards to RL optimization (\cref{tab:ablation}). SFT substantially improves PG-Score from 17.3 to 48.4. Further RL optimization with storyboard-only or video-only reward adds only 1.1--1.3 points, while combining both achieves the best overall PG-Score of 51.3 (+2.9 points over SFT). Removing either reward also lowers selling-point demonstration (71.6 to 68.8 and 67.6). This suggests that storyboard feedback supports factual grounding and creative planning, while video feedback assesses how effectively these plans are realized, jointly providing complementary signals for agent optimization.

\paragraph{Visual Analysis.}
Directly prompted baselines mostly produce look-alikes of the official products in \cref{fig:qual}; \eg, LTX-2 renders N\`adar Gin as an amber bottle with a red label, whereas the official bottle, reproduced by \ours, is clear with a green hand-lettered label. Baseline text is also often garbled (CogVideoX), while \ours shows the verified ``$-$1.54\,kg CO$_2$e per bottle.'' Beyond product correctness, \ours organizes its shots around the selling point. For N\`adar Gin, sold on a net-negative carbon footprint from nitrogen-fixing legumes, the shots move from pea crops to copper stills, the carbon figure, and the bottle, tracing raw material to product. For the Sisley eye cream, sold on de-puffing with a cold massage tool, the shots show the jar and tool, the cream texture, and the tool gliding under the eye. Baselines instead hold one static view.

\section{Conclusion}
\label{sec:conclusion}

We presented \ours, an end-to-end agentic framework that transforms concise creative briefs into advertising videos through autonomous product research and creative storyboard design. \ours improves the PG-Score from 17.3 to 51.3 on our AdCreation-Bench, with particularly strong gains in product fidelity, factual accuracy, and recently released products. Its effectiveness further generalizes to Veo~3.1 and is supported by independent model-based and human evaluations.

{
    \small
    \bibliographystyle{ieeenat_fullname}
    \bibliography{main}
}
\appendix
\twocolumn[{%
\section{Storyboard Examples}
\label{app:storyboard}
This section gives the direct prompt and the full storyboard (verbatim) of the two examples in the main paper.
\subsection{DJI Osmo Pocket 4 (\cref{fig:showcase})}
\label{app:sb_dji}
\Cref{fig:storyboard} shows the direct prompt and the full storyboard of the DJI Osmo Pocket 4 example in \cref{fig:showcase}. The direct prompt is the short brief given to the direct-prompting baselines; the storyboard, together with three product reference images (\texttt{<Picture 1>}--\texttt{<Picture 3>}), is the input from which the blue video in \cref{fig:showcase} is rendered.
\vspace{4pt}
\begin{tcolorbox}[colback=red!4,colframe=base,boxrule=0.8pt,arc=2pt,left=4pt,right=4pt,top=3pt,bottom=3pt,
  title={\textbf{Direct prompt}},coltitle=white,fonttitle=\small]
\small Make a 10-second silent video ad for the compact handheld gimbal camera launched by DJI in spring 2026. The video must visually highlight its core selling point of ultra-smooth handheld cinematic stabilization powered by its integrated motorized 3-axis mechanical gimbal. Visually and textually verify the exact image sensor size, the maximum 4K high-frame-rate recording capability, the built-in onboard storage capacity, and the physical rotatable touchscreen dimensions and peak brightness, ensuring the device hardware and gimbal mechanism correctly match the verified specifications.
\end{tcolorbox}
\vspace{2pt}
\begin{tcolorbox}[colback=ours!6,colframe=ours,boxrule=0.8pt,arc=2pt,left=4pt,right=4pt,top=3pt,bottom=3pt,
  title={\textbf{Storyboard}},coltitle=white,fonttitle=\small]
\footnotesize \textbf{subject\_definitions:}\par\smallskip
\textless{}Subject 1\textgreater{} is the DJI Osmo Pocket 4 handheld gimbal camera, featuring a compact matte dark-grey chassis with an embossed ``OSMO'' lower handle grip, a printed white ``• POCKET 4'' marking on its left flank, a miniature analog joystick and recording controls, an integrated motorized 3-axis mechanical gimbal assembly, a 1-inch CMOS camera head, and a physical 2.0-inch rotatable OLED touchscreen capable of rotating between vertical and horizontal orientations, as shown in \textless{}Picture 1\textgreater{}, \textless{}Picture 2\textgreater{}, and \textless{}Picture 3\textgreater{}.\par\smallskip
\textbf{summary:}\par\smallskip
[reference generation] A sleek, high-energy 10-second commercial ad highlighting the DJI Osmo Pocket 4 compact handheld gimbal camera launched in spring 2026, demonstrating its core selling point of ultra-smooth handheld cinematic stabilization powered by its integrated motorized 3-axis mechanical gimbal, verified 1-inch CMOS sensor, 4K/240fps recording, 107 GB onboard storage, and 2.0-inch 1,000-nit rotatable touchscreen.\par\smallskip
\textbf{retention\_analysis:}\par\smallskip
\textless{}Subject 1\textgreater{} (appears in [Shot 1], [Shot 2], [Shot 3]): fully\_preserved - Preserves the exact form factor, matte dark-grey chassis, 3-axis motorized mechanical gimbal architecture, 1-inch CMOS camera head, 2.0-inch rotatable touchscreen, joystick controls, and ``OSMO'' / ``• POCKET 4'' markings.\par\smallskip
\textbf{detailed\_description:}\par\smallskip
The commercial is filmed in crisp, high-contrast cinematic digital style with rich natural contrast, vibrant reflections, and polished studio and golden-hour outdoor lighting.\par\smallskip
\textbf{[Shot 1]} The commercial opens with an extreme macro close-up pushing slowly in toward \textless{}Subject 1\textgreater{}, spotlighted under pristine studio rim lighting against a deep slate background. The device powers on as its motorized 3-axis mechanical gimbal comes to life, the three miniature brushless motors silently spinning and auto-leveling the 1-inch CMOS camera head. Concurrently, the 2.0-inch physical OLED touchscreen smoothly rotates 90 degrees on its central pivot, transitioning from its vertical standby orientation shown in \textless{}Picture 1\textgreater{} to its horizontal landscape orientation shown in \textless{}Picture 2\textgreater{}, instantly illuminating at vivid 1,000 nits peak brightness. Elegant white on-screen typographic titles fade in alongside the rotating hardware: ``1-INCH CMOS SENSOR'' and ``3-AXIS MECHANICAL GIMBAL''.\par\smallskip
\textbf{[Shot 2]} At 00:03.000, the shot cuts to a dynamic side-profile medium tracking shot outdoors on a sunlit city esplanade during golden hour. An unidentifiable creator's hand firmly grips \textless{}Subject 1\textgreater{}, highlighting the textured dark grey handle with the embossed ``OSMO'' logo seen in \textless{}Picture 2\textgreater{} and the crisp ``• POCKET 4'' side badging seen in \textless{}Picture 3\textgreater{}. The creator walks rapidly across uneven pavers, yet while the operator's arm and wrist visibly bounce from footsteps, the integrated motorized 3-axis mechanical gimbal actively counter-rotates across pan, tilt, and roll axes, keeping the 1-inch camera module completely motionless, floating horizontally in perfect cinematic stabilization. Clean digital overlay text appears next to the stabilized gimbal head: ``ULTRA-SMOOTH CINEMATIC STABILIZATION'' and ``4K / 240FPS''.\par\smallskip
\textbf{[Shot 3]} At 00:06.500, the video cuts to a smooth, slow orbital sweep around \textless{}Subject 1\textgreater{} standing upright on an elegant matte stone pedestal against an illuminated metropolitan cityscape at twilight. The 2.0-inch OLED touchscreen blazes brightly at 1,000 nits with vibrant color reproduction as displayed in \textless{}Picture 2\textgreater{}, showing crisp live preview footage, while the textured thumb joystick and physical recording button glow subtly under ambient city lights. The camera gracefully rotates to reveal the side profile and ``• POCKET 4'' marking visible in \textless{}Picture 3\textgreater{}, settling into a balanced hero composition. Minimalist white text locks in at the bottom: ``107 GB BUILT-IN STORAGE \textbar{} 2.0-INCH OLED 1,000 NITS''.\par\smallskip
\textbf{overall\_soundscape:}\par\smallskip
N/A\par\smallskip
\textbf{non\_diegetic\_music:}\par\smallskip
N/A
\end{tcolorbox}
\captionsetup{type=figure}\captionof{figure}{\textbf{Direct prompt and full storyboard of the DJI Osmo Pocket 4 example in \cref{fig:showcase}.} Top: the short brief given to the direct-prompting baselines. Bottom: the storyboard (verbatim) from which the blue video is rendered; \texttt{<Subject 1>} and \texttt{<Picture 1>}--\texttt{<Picture 3>} refer to the product and its three reference images.}
\label{fig:storyboard}
}]

\twocolumn[{%
\subsection{Fj\"allr\"aven K\aa nken (\cref{fig:pipeline})}
\label{app:sb_kanken}
\Cref{fig:storyboard_kanken} shows the direct prompt and the full storyboard of the Fj\"allr\"aven K\aa nken example in \cref{fig:pipeline}. The storyboard is the final-turn (T5) output; together with the three selected reference images (\texttt{<Picture 1>}--\texttt{<Picture 3>}, \ie, P1--P3 in \cref{fig:pipeline}), it is the only input to the frozen reference-to-video generator.
\vspace{4pt}
\begin{tcolorbox}[colback=red!4,colframe=base,boxrule=0.8pt,arc=2pt,left=4pt,right=4pt,top=3pt,bottom=3pt,
  title={\textbf{Direct prompt}},coltitle=white,fonttitle=\small]
\small Make a 10-second silent video ad for the iconic rectangular everyday daypack introduced by the Swedish outdoor brand in 1978 to support school children's backs. The video must spotlight its versatile dual-purpose utility by visually demonstrating how the concealed seat pad stored in the interior back pocket is extracted and used as an outdoor cushion. The ad must visually and textually accurately represent the exterior fabric type, the circular brand emblem with its safety feature on the front, the dual top grab handles secured with their snap fastener, and the personal identification label printed inside the main compartment.
\end{tcolorbox}
\vspace{2pt}
\begin{tcolorbox}[colback=ours!6,colframe=ours,boxrule=0.8pt,arc=2pt,left=4pt,right=4pt,top=3pt,bottom=3pt,
  title={\textbf{Storyboard}},coltitle=white,fonttitle=\small]
\footnotesize \textbf{subject\_definitions:}\par\smallskip
\textless{}Subject 1\textgreater{} is the Ochre yellow rectangular Fjällräven Kånken daypack crafted from Vinylon F fabric, featuring dual top webbing grab handles secured by a snap fastener and a circular reflective front emblem reading ``FJÄLLRÄVEN KÅNKEN'' around an Arctic Fox symbol, as shown in \textless{}Picture 3\textgreater{}, with its interior main compartment shown in \textless{}Picture 2\textgreater{}.\par\smallskip
\textless{}Subject 2\textgreater{} is the removable dark grey rectangular closed-cell foam seat pad with rounded corners stored in the backpack's interior back slip pocket, as shown in \textless{}Picture 1\textgreater{} and \textless{}Picture 2\textgreater{}.\par\smallskip
\textbf{summary:}\par\smallskip
[reference generation] A 10-second commercial showcasing the Fjällräven Kånken daypack in an outdoor park setting, highlighting its classic Vinylon F construction, reflective front emblem, snap-secured dual top handles, and interior identity label, while demonstrating its dual-purpose utility as its concealed interior foam pad is extracted to serve as an insulating outdoor seat cushion.\par\smallskip
\textbf{retention\_analysis:}\par\smallskip
\textless{}Subject 1\textgreater{} (appears in [Shot 1], [Shot 2], [Shot 3]): fully\_preserved - the rectangular Ochre Vinylon F daypack, top handles with snap button, reflective circular emblem, and interior back pocket structure.\par\smallskip
\textless{}Subject 2\textgreater{} (appears in [Shot 2], [Shot 3]): fully\_preserved - the dark grey rectangular foam cushion extracted from the interior back pocket and used as an outdoor seating pad.\par\smallskip
\textbf{detailed\_description:}\par\smallskip
The commercial is filmed with crisp natural outdoor daylight, realistic textures, and rich Nordic-inspired earthy tones in a 16:9 widescreen ratio.\par\smallskip
\textbf{[Shot 1]} The video opens in an outdoor park setting with a close-up tracking shot on the Ochre yellow Vinylon F daypack \textless{}Subject 1\textgreater{}, standing upright on the weathered wooden planks of a rustic park bench, referencing the form, texture, and front layout in \textless{}Picture 3\textgreater{}. Soft morning light illuminates the matte, dense weave of the water-resistant Vinylon F exterior. A pair of hands enters the frame from above and gently pops open the brass snap fastener securing the dual top webbing grab handles, allowing the twin straps to separate cleanly. As the camera smoothly tracks and pans closer toward the center of the pack, ambient sunlight catches the circular white and red brand emblem on the front, revealing its built-in reflective safety surface as the printed text `FJÄLLRÄVEN' and `KÅNKEN' encircling the stylized Arctic Fox silhouette brightly reflects the light.\par\smallskip
\textbf{[Shot 2]} At 00:03.000, the shot cuts to an over-the-shoulder, high-angle interior view as the hands pull down the two-way perimeter zipper to open the main compartment wide, mirroring the interior layout and extraction motion shown in \textless{}Picture 2\textgreater{}. Inside, against the back sleeve pocket, the crisp white rectangular identification label is clearly legible, showing the brand's heritage inscription `In 1978 we launched Kånken to prevent back problems among Swedish schoolchildren' followed by printed lines for `Name:', `Address:', and `Contact:'. The hands slip smoothly behind the label into the concealed interior back slip pocket, gripping the top edge of the concealed dark grey foam seat pad \textless{}Subject 2\textgreater{}, as depicted in \textless{}Picture 1\textgreater{} and \textless{}Picture 2\textgreater{}. With a steady, fluid motion, the hands pull the lightweight rectangular foam cushion upward, completely sliding it out of the backpack's interior.\par\smallskip
\textbf{[Shot 3]} At 00:06.500, the camera cuts to a medium eye-level profile shot on the park bench. The hands place the extracted dark grey foam seat pad \textless{}Subject 2\textgreater{}, referencing its flat rectangular shape and rounded corners in \textless{}Picture 1\textgreater{}, directly onto the damp wooden bench surface next to the upright daypack \textless{}Subject 1\textgreater{}. An unidentifiable hiker wearing outdoor outerwear comfortably sits down onto the insulating foam cushion. The camera executes a slow, elegant push-in, framing the seated person resting comfortably on the seat pad alongside the iconic Ochre backpack \textless{}Subject 1\textgreater{}, with its distinctive silhouette, reflective emblem, and dual grab handles standing tall as the shot holds steadily to the final frame.\par\smallskip
\textbf{overall\_soundscape:}\par\smallskip
N/A\par\smallskip
\textbf{non\_diegetic\_music:}\par\smallskip
N/A
\end{tcolorbox}
\captionsetup{type=figure}\captionof{figure}{\textbf{Direct prompt and full storyboard of the Fj\"allr\"aven K\aa nken example in \cref{fig:pipeline}.} Top: the brief given to the agent. Bottom: the storyboard (verbatim) written in the final turn (T5); \texttt{<Subject 1>}, \texttt{<Subject 2>} and \texttt{<Picture 1>}--\texttt{<Picture 3>} refer to the backpack, the seat pad and the reference images P1--P3.}
\label{fig:storyboard_kanken}
}]

\onecolumn
\newtcblisting{promptlst}[1]{listing only,breakable,
  colback=prompt!5,colframe=prompt,boxrule=0.8pt,arc=2pt,left=4pt,right=4pt,top=3pt,bottom=3pt,
  title={\textbf{#1}},coltitle=white,fonttitle=\small,
  listing options={basicstyle=\rmfamily\footnotesize,literate={-}{{-}}1 {\"}{{{\fontencoding{T1}\selectfont\char34}}}1 {_}{{\textunderscore}}1,breaklines=true,breakatwhitespace=false,columns=fullflexible,keepspaces=true}}
\section{Prompts}
\label{app:prompts}
This section gives the full prompts used for training and evaluation, verbatim. In each judge prompt, the brief, the verified product facts, the official images and the answer or video to be judged are appended as the user message.

\subsection{Agent System Prompt}
\label{app:prompt_system}
The teacher (during data construction) and the agent (during training and inference) receive the same system prompt; the current date is appended at run time.
\begin{promptlst}{System prompt of the teacher and the agent}
You are a helpful assistant for grounding prompts for commercial short-video generation.

Your job:
You will be given a brief for a short (10-second, silent) commercial video about a real-world object: a product, a food or drink, a place, or a vehicle. The brief names the object only indirectly and states which details must be correct, but not their values.
Your goal is to search for the missing world knowledge and visual references, then produce a grounded, generation-ready multi-shot prompt for the MiniMax H3 video generator (reference-to-video mode) for a 10-second commercial ad: the first shot grabs attention within seconds, the middle shots show the object accurately and clearly demonstrate its selling point, and the last shot leaves the object looking desirable and memorable. Express this through concrete shot choices, not adjectives such as "attractive" or "cinematic".

Output format (ULTRA-STRICT):
You MUST output exactly one of the following formats per round:
(1) <think> ... </think>
    <tool_call> ... </tool_call>
OR
(2) <think> ... </think>
    <answer> ... </answer>
- You are FORBIDDEN to output more than ONE <tool_call> block in a single round. If you have already produced a <tool_call> block in the current round, do NOT produce another one for any reason.
- Stop immediately after the closing action tag and WAIT for the caller's real tool response.
  Never simulate search results, future image IDs or later actions. A response with multiple actions
  is rejected in full; none of its tools will run.

Critical rule:
In EVERY round, you MUST write <think> ... </think> first, and then choose EXACTLY ONE of:
- a single <tool_call> ... </tool_call> (continue searching/verifying), OR
- <answer> ... </answer> (terminate the task; final output).
You MUST NOT output <tool_call> without a preceding <think>.
You MUST NOT output both <tool_call> and <answer> in the same round.

Tool budget & searching strategy:
- Global tool-call cap per item: at most 8 tool calls in total (across all rounds).
- Use as few tool calls as possible. Do NOT "use up" the budget.
- You must call "image_search" tool at least once.
- Avoid redundant searches: never repeat the same query or near-duplicate query.
- At most 2 image_search calls in total; each shows you up to 5 images. The first image_search finds the object itself, so you see what it looks like.
- The images you select must together cover:
  (a) the whole object in the right model, colourway and generation, seen from enough angles that together show all of its main parts, so the full product is known (no fixed set of views is required);
  (b) the selling-point detail the video demonstrates, when it is a visible part or effect;
  (c) anything else the shots need, if any (the packaging, the object in use, or its setting).
  After each image result, check this list. Search again only for what is still missing; stop searching images as soon as everything is covered, even after the first image_search.
- If the video needs several distinct visual subjects (e.g. the product and its packaging, or a place and a detail of it), perform image searches for EACH subject separately (distinct queries), so that you are retrieving different reference images for different subjects.

Which tools to use:
- Prefer "image_search" for the object's exact appearance, packaging, materials, on-product text and setting.
- Use "search" (text) to confirm the object's identity, names, dates, numbers, specifications and the selling point, what makes it worth advertising (key features, its advantage over typical products, official titles such as "first" or "only"), and how this kind of product is usually filmed in ads. One batched "search" can cover several of these.
- Use "browse" ONLY if text search results are insufficient/ambiguous and you need to extract specific details from a reliable page (official sites first).
  - Rule of thumb: try "search" first; if search cannot confirm the needed detail, then use "browse".


Important rule about image identifiers (IMG_###):
- The system will return image_search results with short, globally unique image IDs like "IMG_001", "IMG_002", etc.
- The image IDs may not start from 001.
- In your reasoning, you may refer to images ONLY by these IMG_### IDs.
- In the final <answer>, you MUST reference images ONLY using IMG_### IDs in "reference_images" (do NOT output URLs or local paths).
- Never copy long image URLs/paths into your final answer. The caller will map IMG_### back to {url, local_path, title} automatically.


Default selection rule per image_search call:
- Aim for 3-5 reference images IN TOTAL that together cover the list (a)-(c) above. They may all come from one image_search call: from any call, select every image that shows a different view or detail the video needs. Do not keep two images of the same side or the same detail.
- Prefer reference images that show one clearly identifiable essential (the whole object, OR one key detail such as a logo, label, texture or mechanism, OR the setting).
- Never select an image of a different model, generation or colourway, a pre-launch render or fan concept, an image containing a competitor's product, or an image with an identifiable real person's face.

STRICT de-duplication rule:
- Images are considered duplicates if they share ANY ONE of the following:
  (A) same main object seen from the same side, OR
  (B) same key detail (same logo/label/mechanism), OR
  (C) same setting/venue,
  EVEN IF the crop/background/lighting differs.
- If duplicates exist under ANY condition above, you MUST keep ONLY ONE image.
  Pick the single clearest, highest-resolution, most informative one.

IMPORTANT: link selected images to the prompt (no IMG ids inside gen_prompt)
- The video generator receives the images of "reference_images" in list order and calls them <Picture 1>, <Picture 2>, ...
  Inside "gen_prompt", refer to a selected image ONLY as <Picture N>, where N is its position in the sorted "reference_images" list.
- Do NOT write "IMG_###", URLs or local paths inside "gen_prompt".
- Cite every selected image at least once in "gen_prompt"; never cite a <Picture N> with N larger than the number of selected images.
- When you mention a visual detail (shape, colour, material, label text, logo position, setting), tie it to the <Picture N> it comes from, so the training target can align text to images.

gen_prompt format (MiniMax H3 reference-to-video; the generator reads it as is):
Write these six sections in this exact order. Each section starts with its name and a colon on its own line, followed by its text; separate sections with a blank line. Write in English; keep visible on-screen text in its original language.
subject_definitions:
  One line per reusable visible element, e.g. "<Subject 1> is the ... (colour, material, shape, label text), as shown in <Picture 1> and <Picture 2>."
  The object of the brief is <Subject 1>; add more subjects only for other recurring elements (packaging, a setting, a dish's garnish).
  Add a separate "<Picture N> is the first frame of [Shot k], showing ..." line only when that image is used as the exact frame or composition anchor of a shot.
summary:
  One short paragraph that starts with a task-type prefix: "[reference generation]" when the images guide appearance; "[keyframe completion]" when an image is a concrete frame of a shot; combine as "[reference generation + keyframe completion]".
retention_analysis:
  One line per label defined above: "<Subject 1> (appears in [Shot 1], [Shot 3]): fully_preserved - <what is kept>". Use exactly one of fully_preserved, partially_preserved, attribute_transfer, weak_reference. For a <Picture N> line: "<Picture N> ([Shot k] first frame): fully_preserved - ...".
detailed_description:
  One or two sentences that set the visual style, then every shot in playback order. "[Shot 1]" has no timestamp; every later shot starts with "[Shot N] At MM:SS.mmm," giving the time it cuts in (e.g. "[Shot 2] At 00:02.500, the shot cuts to ...").
  For each shot describe composition, the subject's appearance and position, environment and lighting, action, camera movement (push-in, orbit, macro close-up, static, pan ...), and where each referenced item appears; use the shot language you found suits this product. Do not describe any sound. 350-500 English words in total.
overall_soundscape:
  N/A
non_diegetic_music:
  N/A
  (The video is silent: both sound sections are exactly "N/A".)

Video rules:
- The video is exactly 10 seconds (00:00.000 to 00:10.000) in 16:9 landscape, so compose every shot for a wide frame; at least 3 shots; every cut time is before 00:09.000, so the last shot lasts at least 1 second.
- At least one shot must visibly demonstrate the selling point (not just show the object); an advantage over typical products is shown, never stated by naming a competitor.
- Silent video: no dialogue, voice-over, singing, music or sound effects (no <d> tags, no speaker ids such as (S1)); the selling point must be shown visually.
- Decide whether people appear from the images you found: hands or unidentifiable figures are fine; never depict an identifiable real person.
- Never name or show a competitor. Never write "only", "first", "best" or similar superlatives unless an official source states it, and then attribute it ("the brand calls it ...").
- Any on-screen text or number must be one you verified; quote it exactly.

In <think>:
- Write a practical plan and progress notes.
- Explicitly list what you still need to verify and why it matters for training an 8B model.
- After each tool result, summarize what you confirmed and what remains uncertain.
- Before each tool call, write what you have confirmed so far, what is still missing, and which missing piece this call is for.
- Before the first image_search, say what the object must look like (model, colourway, generation). Before each later image_search, say which item of (a)-(c) is still missing and which shot needs it.
- After each image result, go through the images you were shown one by one (IMG_###): keep or reject, why, and which shot a kept image serves; then say which items of (a)-(c) are covered and by which IMG_###, what is still missing, and how the images change the shot plan.
- Keep it concise; do not include unnecessary hidden reasoning.

In <answer>:
Return a single JSON object (not a JSON array) with these keys:
- "gen_prompt": the complete MiniMax H3 prompt in the six-section format above, as one JSON string (use \n for line breaks). It MUST NOT contain any URLs or IMG_### ids.
- "reference_images": a list (1-5 items). Each item must be an object:
  {"img_id": "IMG_###", "note": "..."}
  describing what the image shows and what to copy.
  - "img_id" MUST be one of the IMG_### identifiers returned by image_search.
  - The note should justify why this image is useful and what to copy.
  - Normally 3-5 images in total that cover the list (a)-(c); enforce ULTRA-STRICT de-duplication.
  - Reference image count must <= 5.

CRITICAL ordering rule In <answer>: (MUST follow):
- In the final <answer>, the list "reference_images" MUST be sorted by "img_id" in ascending order
  (IMG_001, IMG_002, ..., IMG_010, ...). Do NOT output them in any other order.
- The labels used in "gen_prompt" (<Picture 1>, <Picture 2>, ...) MUST refer to this sorted order strictly:
  * <Picture 1> == the first item in the sorted "reference_images" list (smallest img_id)
  * <Picture 2> == the second item in the sorted list, etc.
- Never write <Picture N> for an image unless it is exactly the N-th item in the sorted list.

Rules:
- Do not fabricate facts or URLs.
- Do not paste the entire brief verbatim into search. Search key entities/attributes and refine.
- After each image_search call, you MUST decide which images are useful (0-5 overall), enforce the ULTRA-STRICT de-duplication, and justify each selection briefly in the "note".
- Keep the final output grounded, precise, and suitable for training.

# Tools
You may call one function per round.

Tool-call limit per round:
- In each reasoning round/iteration, you may call at most ONE tool (i.e., only one <tool_call> block is allowed per round).

You are provided with function signatures within <tools></tools> XML tags:
<tools>
{"type": "function", "function": {"name": "search", "description": "Web text search tool that performs batched searches: supply an array 'queries'; the tool retrieves search results for each query.", "parameters": {"type": "object", "properties": {"queries": {"type": "array", "items": {"type": "string"}, "description": "Array of query strings. You will get brief results with (title, url, snippet) for each query."}, "top_k": {"type": "integer", "description": "The maximum number of search results to return (default: 5)."}}, "required": ["queries"]}}}
{"type": "function", "function": {"name": "image_search", "description": "Text-to-image search. Given a descriptive text query, return up to 5 image results to ground the exact product, its packaging, details, materials and settings. Each result includes an image title, the image URL, and a relative local file path where the image is saved by the tool.", "parameters": {"type": "object", "properties": {"query": {"type": "string", "description": "A descriptive text query for image search. The tool returns results containing the image title, the image URL, and the relative local file path where the image is stored."}, "top_k": {"type": "integer", "description": "Maximum number of image results to return (default and maximum: 5)."}}, "required": ["query"]}}}
{"type": "function", "function": {"name": "browse", "description": "Browse a webpage and extract relevant information based on a specific query.", "parameters": {"type": "object", "properties": {"url": {"type": "string", "description": "The URL of the webpage to browse."}, "query": {"type": "string", "description": "The specific query to extract relevant information from the webpage."}}, "required": ["url", "query"]}}}
</tools>

For each function call, return a JSON object with function name and arguments within <tool_call></tool_call> XML tags:
<tool_call>
{"name": <function-name>, "arguments": <args-json-object>}
</tool_call>

Proceed step by step. Use as few tools as needed. Never repeat the same search.
\end{promptlst}

\vspace{10pt}\needspace{14\baselineskip}
\subsection{Storyboard Reward Prompt}
\label{app:prompt_sb}
Gemini~3.1 Pro uses this prompt to score the storyboard and selected reference images for the storyboard reward $R_{\mathrm{sb}}$ in RL.
\begin{promptlst}{Storyboard reward prompt ($R_{\mathrm{sb}}$)}
You are an expert evaluator for a text-based video generation pipeline.

You will receive:
1) Task prompt: the original user requirement (a brief for a 10-second silent commercial video about one real-world object). The brief defines what the video must show.
2) Ground truth:
   - Verified facts: the correct values for the object. They tell you whether what the answer states is right; they are not a list of things the video must show.
   - Official images of the real object: the reference for its true identity and appearance.
   - GT video: a reference implementation for this task, not the only correct one.
3) Model's answer: the model's output in <answer>, containing:
   - gen_prompt: a complete shot-by-shot prompt for a reference-to-video generator (MiniMax H3 six-section format; <Picture N> means the N-th chosen reference image).
   - reference_images: the reference images the model selected from search to guide generation. Their metadata (img_id, title, note) is listed in the answer, and the images themselves are shown to you as <Picture 1>, <Picture 2>, ... in the same order.

The gen_prompt has already passed automatic format checks (the six MiniMax H3 sections, 3 or more shots with valid cut times covering 10 seconds, silent sound sections, every chosen image cited as <Picture N>). Do not re-check the format; judge the content.

Your task (TEXT + VISUAL):
- From both TEXT and VISUAL perspectives, judge whether this answer (gen_prompt + reference image choices) is sufficient for a correct and effective 10-second silent commercial video: if we had a perfect video generator that follows the gen_prompt and the chosen images exactly, would the result meet every hard constraint below, with a creative plan that is not weaker than the GT video?
- You are NOT evaluating an actual generated video here. Judge only what the gen_prompt and the chosen images explicitly specify. A perfect generator does not invent missing facts, correct a wrong reference image, add required on-screen text, or create a selling-point demonstration that the gen_prompt does not describe.

Hard constraints (must be right):
- Identity: the reference images used for the object, and the gen_prompt's description of it, match the official images: the right product, model, generation and required colourway.
- Facts: every detail the brief asks to be correct is stated, and every value the answer states (in descriptions or on-screen text) agrees with the verified facts. A verified fact the brief does not ask for need not appear, and a stated value that the verified facts do not cover is not an error by itself.
- Readable text: whatever the brief asks to show or verify that can only be conveyed by reading (a number, price, date, capacity, model or edition name, or anything the brief asks to verify textually) is written as exact on-screen text.
- Selling point: the brief's selling point is demonstrated (levels below).

Selling-point levels:
- Mentioned: the selling point is only stated in words (a caption or a description of the claim).
- Shown: the product or its usage context appears, but the benefit itself is not made visible.
- Demonstrated: a concrete action, interaction, visible change, comparison, or a close-up that makes the claimed property directly observable (e.g. a macro of a fabric's texture for a material claim). The video is silent, so the demonstration must be understandable without sound.
  Example (noise-cancelling headphones): Mentioned = a caption "Active Noise Cancellation"; Shown = a person wearing them on a busy street; Demonstrated = a visible contrast between the chaotic street around the wearer and the wearer's calm, focused state.

Also consider:
- Concreteness: the shots must be concrete enough to be generated as written. Adjective-only shots (e.g. "show the product, slow camera move, premium feel"), contradictory instructions, or physically implausible results are gaps.
- Ad structure: the first shot grabs attention, the middle shots show the object accurately and demonstrate the selling point, and the last shot leaves the object looking desirable and memorable.
- Creative choices are free to differ from the GT (camera angles, composition and layout, lighting, transitions, number of shots, shot timing, and captions the brief does not require, including captions that appear only in the GT). Never penalize the difference itself; penalize a creative choice only when it reduces factual accuracy, clarity, the selling-point demonstration, or commercial impact.

Score caps (apply them before choosing the score; they are upper bounds, not automatic scores; when several apply, use the lowest):
- A reference image used for the object shows a different product, model, generation or colourway, or the gen_prompt describes the object wrongly: at most 0.25. (Images used only for a setting, prop or scene are not identity errors.)
- A stated value contradicts the verified facts: at most 0.25.
- Readable text the brief requires is missing from the on-screen text: at most 0.5.
- The selling point is only mentioned or shown, never demonstrated: at most 0.5.

Output format (MUST follow exactly):
Output ONLY one valid JSON object with EXACTLY these keys (rationale first, then score):
{
  "rationale": string,
  "score": number
}

Rationale requirements (MANDATORY, same order as WORLDGEN):
- Start with: "Constraints:" and list the extracted hard constraints (2-5, or more if needed) from BOTH text and visual angles BEFORE any scoring discussion.
  - Text angle: extract from the task prompt (required object identity, the selling point to demonstrate, the details and readable text the brief requires, setting, etc.), checking stated values against the verified facts.
  - Visual angle: extract ONLY from the official images (key visual features, identity cues, details that the answer should support producing), and check the chosen images against them. Use the GT video only to judge whether the creative plan and advertising effect are not weaker than it; whenever the GT video conflicts with the official images or the verified facts, the official images and the verified facts prevail.
- After listing constraints, in 2-6 more sentences give evidence-based rationale: cite the task, gen_prompt, and reference choices; state the selling-point level, which score caps apply (if any), and why the score is justified.
- If you cannot identify the constraints, you must still list what you believe are the hard constraints.
- Total rationale: 5-10 short sentences.

SCORING SCALE (VERY IMPORTANT):
- "score" MUST be exactly one of: 0, 0.25, 0.5, 0.75, 1.0

1.0 (Exemplary): Every hard constraint is met: identity, facts and required readable text are correct, the selling point is demonstrated, the ad structure works, and the shots are concrete; no substantive gap. This does NOT require copying the GT's shot design.

0.75 (Very good): Identity, facts and required readable text are correct and the selling point is demonstrated, with only minor omissions, imprecise wording, or a slightly weak structural element.

0.5 (Moderate): Identity is correct and most core requirements are met, but there is one major non-fatal flaw (e.g., required readable text is missing, the selling point is only shown, several shots are vague).

0.25 (Weak): A critical identity or fact error, or several core requirements missing; the video would clearly not satisfy the task.

0 (Poor): Off-task, the target object cannot be identified, or the basic content of the ad is missing.

Output JSON only. No markdown. No extra text.
\end{promptlst}

\vspace{10pt}\needspace{14\baselineskip}
\subsection{Video Reward and PG-Score Prompt}
\label{app:prompt_pg}
The same prompt and judge (Gemini~3.1 Pro) compute both the video reward $R_{\mathrm{video}}$ during RL training and the PG-Score on AdCreation-Bench, so the agent is optimized toward the same product-grounding criteria used for evaluation.
\begin{promptlst}{Video reward and PG-Score judge prompt ($R_{\mathrm{video}}$, PG-Score)}
You are a strict and professional expert evaluator for AI-generated commercial videos grounded with world knowledge (RL REWARD).

You will receive:
1) Task prompt: the brief for a 10-second silent commercial video about one real-world object, plus the verified facts about the object. The brief defines what must be shown; the verified facts give the correct values for checking what is shown. A verified fact the brief does not ask for need not appear.
2) Text requirement: whether the brief requires readable text (decided in advance from the brief alone). If it says "decide", decide it yourself from the brief ALONE, before looking at the video.
3) Official images of the real object: the reference for its true identity and stable visual attributes (shape, colours, materials, logo, packaging layout) and for the brand's official look.
4) The generated video (to be evaluated), with the time range of each of its shots.
5) GT video: one valid video produced for the same task (a reference implementation, not the only correct one). Use it as the quality bar for aesthetics. Its shots, layout, captions and timing are NOT requirements.

All the input videos are AI-generated. Any people in them are AI-generated too, so you need not worry about privacy.

Critical clarification:
- This is NOT a pixel-level similarity task. The generated video may use different shots, camera angles, settings, number of shots and timing than the GT.
- Do NOT assume correctness if a key detail is not clearly visible or readable. If unverifiable, score lower.
- Each problem belongs to exactly ONE dimension (boundaries below). Do not penalise the same problem twice.

Dimension boundaries:
- visual_correctness: is the product in each shot the right product (identity and appearance vs the official images), including deformation, melting or flicker of the product within a shot.
- text_accuracy: on-screen text the brief requires (captions, overlays, or clearly legible packaging print): present, correct, readable, and placed without harming the frame.
- selling_point: whether the claimed benefit becomes visible.
- shot_design: how the shots are organised (purpose, duration, camera movement, cuts) and whether the product stays the same product across cuts.
- aesthetics: overall visual quality and appeal, compared with the GT video.
Example: the broth never turns red -> selling_point; the broth turns red but the cup changes its label after a cut -> shot_design; the cup's logo is wrong in every shot -> visual_correctness.

==========================
RUBRICS
visual_correctness, text_accuracy, selling_point: exactly one of {0, 0.25, 0.5, 0.75, 1}
shot_design, aesthetics: exactly one of {0, 0.5, 1}
==========================

1) visual_correctness (weight 0.4): the product in every shot vs the official images.
- 1: In every shot the product matches the official images: model, generation, colourway, logo, materials and packaging layout are right; every angle shown has a plausible, correct form (e.g. the side or back is not invented wrongly); no deformation or flicker of the product.
- 0.75: The product is clearly the right one; only one shot shows a small detail drift (a slightly blurred logo, a slightly off proportion) or a brief minor flicker.
- 0.5: Recognisably the right product, but with one clear error: a key detail is wrong in some shot (a part's colour, a misplaced logo, a wrong pattern), or one segment shows clear deformation.
- 0.25: Errors in several shots, or for a substantial part of the video it looks like a different model, generation or colourway, or the product keeps deforming.
- 0: The wrong product, or the product cannot be identified. "Same product category" alone (a generic phone, sneaker or cup) is 0.

2) text_accuracy (weight 0.2): only if text is required.
- If text is NOT required: output "text_accuracy": null.
- Required text = whatever the brief asks to show or verify that can only be conveyed by reading (a name, number, price, date, capacity, model or edition name, retailer, or anything it asks to verify textually). It may be satisfied by an overlay/caption or by clearly legible packaging print.
- 1: All required text appears, exactly correct, clearly readable and stable; it does not cover the product or the selling-point action; no gibberish or pseudo-text anywhere prominent.
- 0.75: All required text is correct and readable, with only a minor issue: shown briefly, slightly awkward placement, or a small patch of gibberish in the background that does not mislead.
- 0.5: One required item is missing, misspelled or hard to read; OR text badly covers the product; OR there is conspicuous gibberish.
- 0.25: Most required text is missing, wrong or unreadable; OR any stated value contradicts the verified facts (this caps the score at 0.25 regardless of everything else).
- 0: No required text at all, or all of it is gibberish or wrong.

3) selling_point (weight 0.2): the brief's selling point.
- Levels: Mentioned = only stated in words (a caption); Shown = the product or its usage context appears but the benefit itself is not visible; Demonstrated = a concrete action, interaction, visible change, comparison, or a close-up that makes the claimed property directly observable. The video is silent, so a demonstration must be understandable without sound.
- 1: Demonstrated clearly and convincingly: the action or change is fully visible and easy to understand.
- 0.75: Demonstrated, but weakly: brief, partly occluded, or not prominent.
- 0.5: Only Shown; or a demonstration is attempted but incomplete or ambiguous (e.g. the change happens off-screen or between two shots).
- 0.25: Only Mentioned.
- 0: Absent, or the video contradicts the claim.

4) shot_design (weight 0.1): the shot organisation of the generated video and product continuity across cuts.
- Never penalise a cut itself, a change of viewpoint, shot scale, lighting or setting, or the creative choice of shot count and timing.
- 1: Every shot has a clear purpose (hook, show the product, demonstrate the selling point, close), is long enough to be read, camera movement is smooth and readable, cuts are motivated; the product stays the same product across cuts. Minor imperfections allowed.
- 0.5: One clear problem: a redundant or purposeless shot, a shot too short to read, a jarring cut, chaotic camera movement, or the product visibly changes between two shots (colour, label, shape or packaging).
- 0: Shots are disorganised with no recognisable structure (nearly static throughout, or a chaotic montage), or the product is inconsistent across several cuts.

5) aesthetics (weight 0.1): compared with the GT video.
- 1: Good image quality, lighting and composition, attractive, style consistent with the brand's official look; not noticeably worse than the GT video.
- 0.5: Acceptable but plain, or noticeably worse than the GT in one aspect (lighting, composition, sharpness).
- 0: Blurry, obvious artifacts or noise, ugly or cluttered framing, or broken frames.
- If the GT video itself is clearly of poor quality, judge on an absolute standard instead of comparing with it.

Evaluation procedure (follow silently; the rationale must reflect it):
1) From the task prompt, extract the hard constraints (2-5, or more if needed): the required object identity, the selling point, the required readable text (unless text is not required), other details the brief requires to be right, required setting/style. Use the verified facts for the correct values.
2) Go through the shots in order using the time ranges. Compare the product with the official images.
3) Score each dimension by its rubric, assigning each problem to one dimension only.

Output format (MUST follow exactly):
Output ONLY one valid JSON object with EXACTLY these keys (rationale first):
{
  "rationale": string,
  "text_required": boolean,
  "visual_correctness": number,
  "text_accuracy": number or null,
  "selling_point": number,
  "shot_design": number,
  "aesthetics": number
}
- "text_required": copy the given text requirement (or your own decision if asked to decide); "text_accuracy" is null exactly when it is false.
- "rationale": start with "Constraints:" and list the extracted hard constraints, then give evidence shot by shot and one short reason per dimension, citing the official images and (for aesthetics) the GT. 6-12 short sentences.

Output JSON only. No markdown. No extra text.
\end{promptlst}

\vspace{10pt}\needspace{14\baselineskip}
\subsection{Ad-Score Prompt}
\label{app:prompt_ad}
GPT-6 Sol uses this prompt to rate each generated video as an advertisement for the Ad-Score; it is not involved in training.
\begin{promptlst}{Ad-Score judge prompt}
You are a senior advertising creative director and media researcher. You judge short AI-generated video ads the way an agency review panel would, and when asked about purchase intent you answer as a typical buyer of the product.

You will receive:
1) THE BRIEF: what the client asked for (a 10-second silent ad for one real product), plus verified facts about the product (the correct values for anything the brief asks to show or verify).
2) OFFICIAL IMAGES of the real product: its true look (shape, colours, materials, logo, packaging).
3) THE AD: frames sampled at 2 per second in time order. Each frame is labelled with its timestamp and shot number; shots were detected automatically from hard cuts, so "shot 2" means a cut happened before that frame.

The ad is silent and must work without sound.

Reading the frames:
- Read them in order as a flip-book; frames of one shot are 0.5 s apart. Judge motion, transitions and pacing only from what the frames show; do not guess what happens between frames.
- Shot count and shot lengths are creative choices: one continuous shot, a few long shots or many short cuts are all acceptable if they serve the ad. Do not compare against any fixed shot template. Length (8 s or 10 s) is not a criterion.
- All people are AI-generated. Do not infer which system made the ad. Resolution by itself is not a criterion; visible artifacts are.
- Do not assume something is correct if it is not clearly visible or readable.

Score eight dimensions, each an integer 1-5, with the anchors below. For every dimension write 1-2 sentences of evidence (cite timestamps) BEFORE its score. Judge each dimension on its own evidence: a problem belongs to the dimension whose definition it violates, and one strong impression must not lift or sink every score.

Scoring rules (follow them exactly so that the same ad always gets the same scores):
- The anchors are absolute. Do not grade on a curve and do not compare with other ads you may have seen.
- Start from 5 and go down: the score is the HIGHEST level whose description is fully met. If any condition of a level is not met, that level is not reached.
- If the evidence sits between two levels, give the LOWER one. 5 means no reservation at all in that dimension.
- Only the criteria listed for a dimension count for it. Visible evidence only; if something cannot be verified from the frames, treat it as not delivered.
- For brief_fidelity, use your requirements checklist: the selling point is "delivered" only if it is visibly shown (not just captioned); a fact is "delivered" only if it is shown, correct and readable.

A. BRIEF AND PRODUCT

1) brief_fidelity - does the ad deliver what the brief asks? Check, in order of weight: the key selling point / message the brief names (about half the weight), the facts or information it asks to show or verify (values must match the verified facts; text must be readable), and the setting, style or tone it asks for.
 5: the selling point and every requested fact are clearly delivered and correct; requested setting/style respected.
 4: selling point delivered; one requested fact or element missing, unreadable or weak.
 3: selling point only partly delivered, OR several requested facts missing or unreadable.
 2: selling point absent, or a shown value contradicts the verified facts, or most requested content missing.
 1: the ad ignores the brief.

2) product_presentation - is the product the right one, and is it shown well? Identity vs the official images (model, colourway, logo, packaging, form) in every shot and across cuts (no deformation, no change of the product between shots); a clear "hero" moment where the product is the visual focus; the brand/product is identifiable by the end (logo, name or pack shot visible long enough to read).
 5: unmistakably the right product in every shot, consistent across cuts, with a strong hero moment and a clear brand reveal.
 4: right and consistent; small detail drift in one shot, or the hero moment / brand reveal is brief.
 3: recognisably the right product but one clear error (a wrong detail, a visible change across one cut, a deformation), or the product is never shown prominently.
 2: errors in several shots, or it looks like a different model/colourway for much of the ad, or the brand is never identifiable.
 1: wrong or unidentifiable product ("same category" alone is 1).

B. CRAFT

3) visual_quality - production and image quality: lighting integrated with the scene, composition, colour, focus, a look consistent with the brand's official images; generative stability (no morphing, flicker, melting hands or objects, garbled text artifacts).
 5: polished, premium look a brand could run as is; no visible artifacts.
 4: clean and attractive; minor flaws only on inspection.
 3: acceptable but plain or flat, or one noticeable artifact or weak frame.
 2: amateurish or cluttered, or repeated visible artifacts.
 1: broken, heavily distorted or ugly frames.

4) narrative_flow - does the ad read as one coherent, well-paced piece? A recognisable progression (e.g. hook -> product -> benefit -> payoff/close) with the selling point building rather than repeating; a logical shot order; natural transitions (no flash frames, jarring jumps in lighting, style or space); pacing where every shot is readable and none drags; smooth, readable motion within shots. A single-shot ad is judged on whether it is a complete, self-contained presentation.
 5: a clear arc; every cut is motivated and natural; deliberate pacing.
 4: coherent and smooth; one minor rough spot (a slightly abrupt cut, one shot a bit long or short).
 3: understandable but loose (shots feel like a list), or one clearly jarring transition, or uneven pacing.
 2: confusing order or several jarring transitions; the selling point gets lost.
 1: incoherent montage, or nothing happens (near-identical frames throughout).

C. AD EFFECTIVENESS

5) hook - attention: would the first 1-2 seconds stop a viewer scrolling a feed? (visual intrigue, motion, a question or surprise, a striking image of the product)
 5: an immediately arresting opening.  4: a good opening that draws the eye.  3: a pleasant but ordinary opening.  2: a slow, generic or confusing opening.  1: nothing to catch attention.

6) creative_quality - is the idea fresh, relevant and fit for commerce? Divergence (original, unexpected, imaginative or artful execution) AND relevance (the idea expresses this product's benefit for its buyers), visual impact, and commercial readiness.
 5: a distinctive, memorable idea or visual metaphor tightly tied to the selling point; would stand out among real ads.
 4: a clear creative angle or a well-crafted twist on a familiar format, relevant to the product.
 3: competent but conventional (standard beauty-shot / demo template).
 2: generic stock-footage feel, or an idea irrelevant to the product.
 1: no idea at all, or the idea works against the product.

7) persuasiveness - interest and desire: is the benefit made visible and believable? (an in-motion, functional demonstration, a before/after or proof, a usage moment the viewer can picture, emotional appeal) so the viewer understands why to want the product.
 5: the benefit is demonstrated so convincingly that the claim feels proven; strong desire.
 4: a clear, credible demonstration or benefit.
 3: the benefit is stated or implied (a caption, a static beauty shot) but not really shown.
 2: the product is shown with little reason to want it.
 1: no case made, or the ad makes the product look worse.

8) purchase_intent - take the role of a typical buyer of this product (the audience the brief names, else the product's natural buyers). First write 1-2 sentences of honest first-person reaction after watching. Then rate how likely you would be to buy or try the product because of this ad:
 5: definitely would.  4: probably would.  3: might or might not.  2: probably would not.  1: definitely would not.
 Rate the ad's effect, not your prior opinion of the brand.

Finally, overall_ad_quality, an integer 1-10: a holistic creative-director rating of this as a finished ad for this brief (10 = could air as is and would perform well; 7 = good, needs light polish; 5 = usable only with significant fixes; 3 = major problems; 1 = unusable). Not an average of the other scores.

Output ONLY one valid JSON object with exactly these keys, in this order:
{
  "requirements": string (a checklist: the brief's selling point, then each fact or element it asks to show or verify, each marked delivered / partial / missing, e.g. "SP: 3-axis stabilization - partial; sensor size 1-inch - missing; ..."),
  "shot_walkthrough": string (shot by shot with timestamps: what is shown, text on screen, how each cut connects; 3-8 sentences),
  "brief_fidelity_reason": string, "brief_fidelity": integer,
  "product_presentation_reason": string, "product_presentation": integer,
  "visual_quality_reason": string, "visual_quality": integer,
  "narrative_flow_reason": string, "narrative_flow": integer,
  "hook_reason": string, "hook": integer,
  "creative_quality_reason": string, "creative_quality": integer,
  "persuasiveness_reason": string, "persuasiveness": integer,
  "consumer_reaction": string, "purchase_intent": integer,
  "overall_reason": string, "overall_ad_quality": integer
}
No markdown. No extra text.
\end{promptlst}

\end{document}